%% file: main.tex
\documentclass[lettersize,journal]{IEEEtran}
\usepackage{amsmath,amsfonts}
\usepackage{amssymb}
\usepackage{algorithmic}
\usepackage{algorithm}
\usepackage{array}
\usepackage[caption=false,font=normalsize,labelfont=sf,textfont=sf]{subfig}
\usepackage{textcomp}
\usepackage{stfloats}
\usepackage{url}
\usepackage{verbatim}
\usepackage{graphicx}
\usepackage{cite}
\usepackage[pagebackref,breaklinks,colorlinks,citecolor=blue]{hyperref}
\usepackage{inconsolata}
\usepackage{caption} 
\usepackage{enumitem}

\usepackage[table]{xcolor}
\usepackage{microtype}
\usepackage{array,multirow,textcomp}
\usepackage{pifont}
\newcommand{\cmark}{\ding{51}}%
\newcommand{\xmark}{\ding{55}}%
\usepackage{bm}
\usepackage{supertabular,booktabs}
\usepackage{makecell}
\usepackage{arydshln}
\usepackage{ulem}
\usepackage{graphicx}
\usepackage{etoolbox}
\newcommand{\insertfig}{\includegraphics[width=\linewidth]{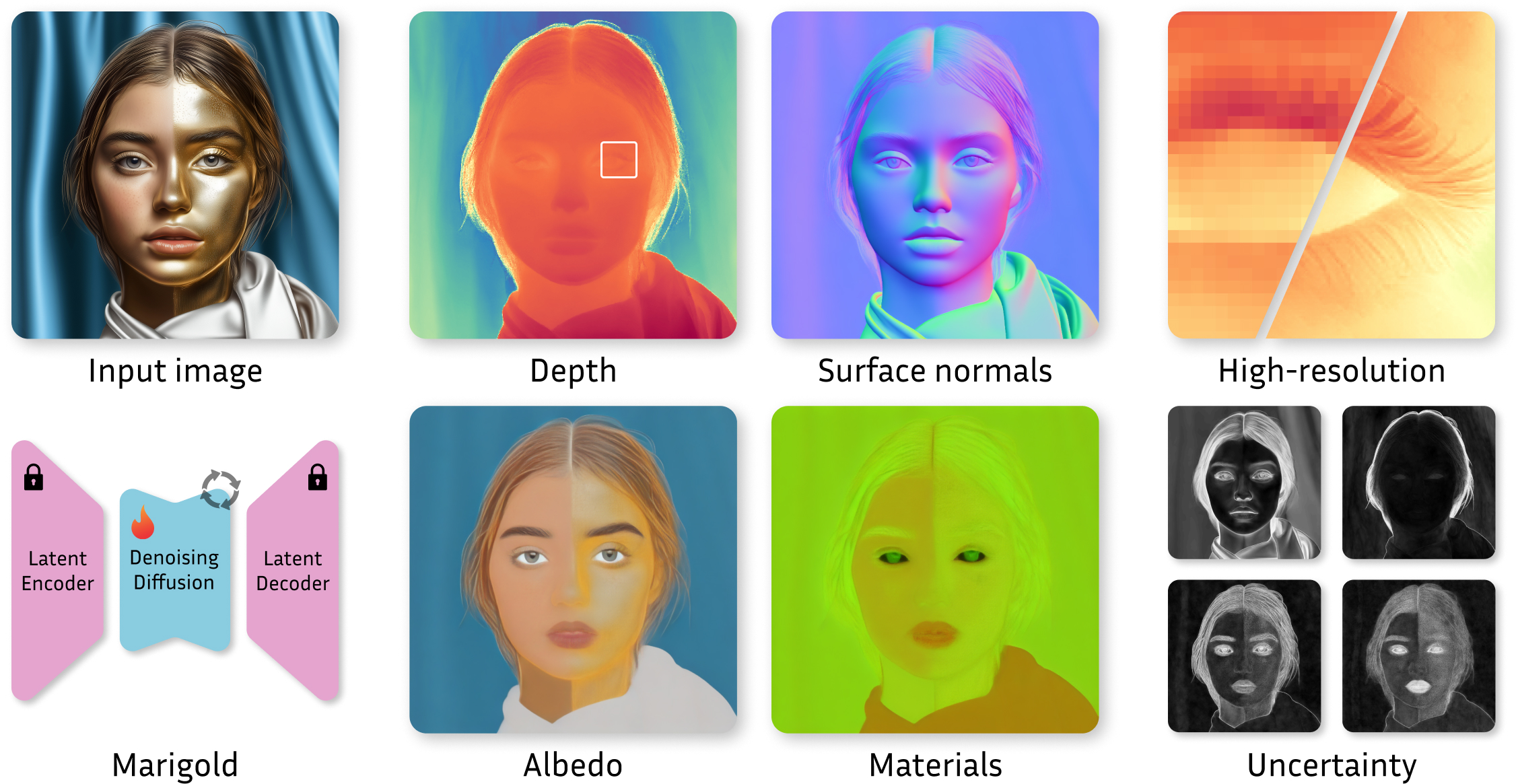}\captionof*{figure}{
    \small
    We present \method{}, a fine-tuning protocol for various image analysis tasks, and a family of associated diffusion models.
    Without loss of generality, these include monocular depth estimation, surface normals prediction, and intrinsic image decomposition.
    Its core principle is to leverage the rich visual knowledge stored in modern generative image models. 
    As a generative model derived from Stable Diffusion and fine-tuned with synthetic data, \method{} can zero-shot transfer to unseen datasets, offering state-of-the-art results.
    The visualizations above demonstrate the strong out-of-distribution performance: without observing a single image other than synthetic rooms and dashboard views, \method{} can extract pixel-perfect depth maps, surface normals, and intrinsic decomposition of images, ready for downstream tasks.
}}
\makeatletter
\apptocmd{\@maketitle}{\centering\vspace{1em}\insertfig\vspace{-1em}}{}{}
\makeatother

\usepackage{xspace}
\makeatletter
\DeclareRobustCommand\onedot{\futurelet\@let@token\@onedot}
\def\@onedot{\ifx\@let@token.\else.\null\fi\xspace}
\def\eg{\textit{e.g}\onedot} 
\def\ie{\textit{i.e}\onedot} 
\def\cf{\textit{cf}\onedot} 
 
\def\wrt{w.r.t\onedot} 
 
\def\etal{\textit{et\,al}\onedot}
\makeatother

\newcommand{\method}{Marigold}%

\newcommand{\nocompete}[1]{\textcolor{gray}{#1}}

\newcommand{\epscompdbe}{\epsilon^{\scriptscriptstyle \text{comp}}_{\scriptscriptstyle \text{DBE}}}
\newcommand{\epsaccdbe}{\epsilon^{\scriptscriptstyle \text{acc}}_{\scriptscriptstyle \text{DBE}}}
\newcommand{\edgeprc}{\epsilon^{\scriptscriptstyle \text{prc}}}
\newcommand{\edgerec}{\epsilon^{\scriptscriptstyle \text{rec}}}

\begin{document}

\title{
Marigold: Affordable Adaptation of Diffusion-Based \\
Image Generators for Image Analysis
}

\author{
    Bingxin Ke$^{*}$, 
    Kevin Qu$^{*}$, 
    Tianfu Wang$^{*}$, 
    Nando Metzger$^{*}$, 
    Shengyu Huang, 
    Bo Li, \\
    Anton Obukhov$^{\dagger,*}$, 
    Konrad Schindler$^{\dagger}$%
    \thanks{
      Work done at the Photogrammetry and Remote Sensing Laboratory, ETH Zürich, Switzerland.
      * denotes equal technical contribution.
      $\dagger$ denotes equal supervision.
      Corresponding author: Konrad Schindler (schindler@ethz.ch).
    }
}

\markboth{IEEE TRANSACTIONS ON PATTERN ANALYSIS AND MACHINE INTELLIGENCE}%
{Ke \MakeLowercase{\textit{et al.}}: Marigold Image Analysis}

\IEEEpubid{0000--0000/00\$00.00~\copyright~2021 IEEE}

\maketitle

\input{sec/0_abstract}
\input{sec/1_intro}
\input{sec/2_related_work}
\input{sec/3_method}

\input{sec/5_conclusion}
\input{sec/ack}

{
    \bibliographystyle{IEEEtran}
    \bibliography{main}
}

\vspace{5em}

\input{sec/bibliography}

\vfill
\end{document}

%% file: sec/0_abstract.tex
\begin{abstract}
The success of deep learning in computer vision over the past decade has hinged on large labeled datasets and strong pretrained models. 
In data-scarce settings, the quality of these pretrained models becomes crucial for effective transfer learning.
Image classification and self-supervised learning have traditionally been the primary methods for pretraining CNNs and transformer-based architectures. 
Recently, the rise of text-to-image generative models, particularly those using denoising diffusion in a latent space, has introduced a new class of foundational models trained on massive, captioned image datasets. 
These models’ ability to generate realistic images of unseen content suggests they possess a deep understanding of the visual world.
In this work, we present \emph{\method{}}, a family of conditional generative models and a fine-tuning protocol that extracts the knowledge from pretrained latent diffusion models like Stable Diffusion and adapts them for dense image analysis tasks, including monocular depth estimation, surface normals prediction, and intrinsic decomposition. 
\method{} requires minimal modification of the pre-trained latent diffusion model's architecture, trains with small synthetic datasets on a single GPU over a few days, and demonstrates state-of-the-art zero-shot generalization.
Project page: \href{https://marigoldcomputervision.github.io}{https://marigoldcomputervision.github.io}.
\end{abstract}

\begin{IEEEkeywords}
Denoising diffusion, image analysis, image generation, foundational models, transfer learning.
\end{IEEEkeywords}

%% file: sec/1_intro.tex
\section{Introduction}
\label{sec:intro}

\IEEEPARstart{T}{he}
introduction of ImageNet~\cite{deng2009imagenet} laid the foundation for training deep Convolutional Neural Networks (CNNs), such as AlexNet~\cite{krizhevsky2012imagenet}, catalyzing further advances in the computer vision field: in data acquisition, neural architectures, and training techniques. 
With the advent of VGG~\cite{simonyan2014very} and ResNet~\cite{he2016deep} architectures, transfer learning~\cite{yosinski2014transferable} became essential for training high-performance computer vision models and reducing training time of semantic segmentation~\cite{long2015fully}, depth prediction~\cite{eigen_depth_2014}, and other downstream tasks.
\IEEEpubidadjcol %
In many cases, training a neural network from random weight initialization is claimed not feasible~\cite{long2015fully}.
Modern deep learning frameworks~\cite{paszke2019pytorch} have since made it easy to use pretrained models by allowing practitioners to load pretrained weights with a simple setting like \texttt{pretrained=True} during model creation.

The rise of large text-to-image generative models\cite{ramesh2021dalle} and denoising diffusion approaches\cite{song2019generative,pmlr-v162-nichol22a} has opened new opportunities for leveraging the rich priors embedded in foundational models.
A breakthrough in this area came with the introduction of Latent Diffusion Models (LDMs), a class of models exemplified by the widely known Stable Diffusion (SD)~\cite{rombach2022high}.
These models operate in the compressed latent space of a pretrained Variational Autoencoder (VAE), enabling significant resource savings in both training and inference. 
Trained on the internet-scale LAION-5B dataset of captioned images\cite{schuhmann2022laion5b}, Stable Diffusion excels in realism and diversity. 
Its open-source availability, low computational requirements for inference, and integration with toolkits like \texttt{diffusers}~\cite{von-platen-etal-2022-diffusers} have enabled widespread experimentation by researchers and artists.
The abundance of customization recipes~\cite{brooks2023instructpix2pix,zhang2023adding,ruiz2023dreambooth} has prompted many notable extensions that focus on enhancing the controllability of the original image generation task.

Repurposing text-to-image LDMs from image generation to image analysis is a recent development in generative imaging. 
The motivation is simple: if a diffusion model demonstrates a deep understanding of the visual world through high-quality image generation, that same understanding can be leveraged to derive a versatile regression model for image analysis.
To this end, in our recent work~\cite{ke2023repurposing}, we introduced \textbf{\method{}-Depth}, an LDM-based state-of-the-art zero-shot affine-invariant monocular depth estimator, along with a simple and resource-efficient fine-tuning protocol for Stable Diffusion. 

\method{}-Depth proposed several key novelties unlocking the potential of LDMs for image understanding: 
\begin{enumerate}[label=(\arabic*)]
\item reusing the LDM's VAE to encode not just the input image but also the output modality into the latent space;
\item using only high-quality synthetic data;
\item short resource-efficient fine-tuning protocol;
\item generative modeling of a conditional distribution rather than predicting its mode as end-to-end approaches do.
\end{enumerate}

Some of these properties are organically entangled. 
(1$\leftrightarrow$2):~Encoding the modality into latent space is only possible when it is noise-free and pixel-complete -- rarely the case with the real depth ground truth.
(2$\leftrightarrow$3):~A short fine-tuning protocol preserves prior knowledge. It requires diverse, consistently labeled, and noise-free data to reduce noise in weight updates, which are satisfiable with synthetic data.
(1$\leftrightarrow$3):~Operation in latent space ensures affordable fine-tuning and inference on a single consumer Graphics Processing Unit (GPU), empowering research even outside large labs.

The importance of synthetic data and strong prior for depth estimation have been subsequently confirmed in Depth Anything V2~\cite{yang2024depthv2}.
Although their end-to-end model achieves impressive performance in zero-shot benchmarks, it involves a 3-stage training procedure, a teacher-student separation, and generating 62M pseudo labels; both do not fit the bill of a simple and affordable transfer learning recipe.

For the property (4), as shown in~\cite{ke2023repurposing}, modeling the distribution of depth conditioned on the input image with an LDM allows for multiple plausible interpretations of the input. 
This ability is essential for solving ill-posed problems, as there may be no single correct output due to over-exposed input, blur, ambiguities of transparent objects, \textit{etc}.
Obtaining samples from the conditional distribution with \method{} is as simple as starting the diffusion process from different noise samples given the same input.
Multiple such predictions can be ensembled to approximate the mode of a conditional distribution.
Ensembling is required to evaluate prediction quality in standard benchmarks comprised of image-depth ground truth pairings, an established evaluation protocol~\cite{Geiger2013IJRR,SilbermanECCV12nyu}.
Additionally, computing predictive uncertainty becomes tractable given such an ensemble.

The multi-step inference and the computational redundancy of the ensembling typically result in many function evaluations (NFEs), which slow down inference speed -- initially, a major point of criticism of the original \method{}-Depth~\cite{ke2023repurposing}.
To this end, we explored Latent Consistency Distillation~\cite{luo2023latent} to reduce the number of sampling steps arbitrarily low, even to just one.
Simultaneously, several works explored other techniques to bring sampling steps down.
Some preserved the generative nature of the model~\cite{garcia2024fine, gui2024depthfm, he2024lotus}. 
Additionally, \cite{garcia2024fine} and \cite{xu2024genpercept} showed the possibility of scoring high in the said benchmarks by re-casting \method{} as an end-to-end network, effectively yielding just one function evaluation.
In this paper, we continue focusing on the generative formulation.

Notably, Garcia~\textit{et al.}~\cite{garcia2024fine} discovered sub-optimal settings in the diffusion scheduler of the original \method{}-Depth~\cite{ke2023repurposing} and proposed a correction, leading to significant improvement of that exact model's performance in the same benchmarks in the few-step inference regime. 
We gratefully incorporated this observation and integrated this regime into our study.
With single-step inference and technical enhancements such as using a lightweight compatible VAE~\cite{taesd} and low-precision weight quantization, \method{} can now produce predictions in under 100ms on most commodity hardware.

In this paper, we follow up on the initial body of work~\cite{ke2023repurposing}, recap it in Sec.~\ref{sec:method}, and extend our study with several dense image analysis tasks~\cite{horn1970shape} having a long history in computer graphics: surface normals prediction in Sec.~\ref{sec:method:normals} and intrinsic image decomposition in Sec.~\ref{sec:method:iid}.
Additionally, we introduce a protocol for distilling \method{} models into Latent Consistency Models (LCMs) in Sec.~\ref{sec:method:lcm} and the new High-Resolution (HR) model and inference strategy in Sec.~\ref{sec:method:hr}.

To summarize, our contributions are:
\begin{enumerate}[label=(\arabic*)]
    \item \textit{\method{}} -- a simple and resource-efficient \emph{fine-tuning protocol} to convert a foundational LDM image generator into a \emph{zero-shot} generative image analysis model with robust in-the-wild generalization capability. 
    Training a \method{} takes less than \emph{3 GPU-days} and can be accomplished with most commodity hardware;
    \item A comprehensive study of training diffusion models with \method{} for \emph{monocular depth estimation}, \emph{surface normal prediction}, and \emph{intrinsic image decomposition} tasks;
    \item \emph{Fast} (sub-100ms) few- or single-step \emph{inference};
    \item Overcoming the resolution bias of the base model, enabling \emph{high-resolution inference}. 
\end{enumerate}

%% file: sec/2_related_work.tex
\section{Related Work}
\label{sec:related_work}

\subsection{Diffusion Models}

Denoising Diffusion Probabilistic Models (DDPMs)~\cite{song2019generative,ho2020denoising_ddpm} generate data by reversing a Gaussian noise diffusion process. 
Denoising Diffusion Implicit Models (DDIMs)~\cite{song2020denoising_ddim} extend this by introducing a non-Markovian shortcut for faster sampling at inference time.
Latent Consistency Models (LCMs)~\cite{song2023consistency,luo2023latent} distill DDPM models into consistency functions that map points on the diffusion ODE trajectory~\cite{song2020score} to the same output, thereby introducing a different parameterization and reducing inference time. 
In the realm of image generation, Rombach~\etal~\cite{rombach2022high} have revolutionized generative modeling with their Stable Diffusion model, trained with LAION-5B~\cite{schuhmann2022laion5b} dataset of 2.3B text-image pairs.
The cornerstone of their approach is an LDM, where the denoising process is run in an efficient latent space, drastically reducing the complexity of the learned mapping.
This model holds rich visual priors, providing a solid foundation for our generalizable image analysis models and a base for our transfer learning approach.

\subsection{Foundation Models}
Vision Foundation Models (VFMs) are large neural networks trained on internet-scale data.
The extreme scaling leads to the emergence of high-level visual understanding, such that the model can then be used as is~\cite{wang2023breathing} or fine-tuned to a wide range of downstream tasks with minimal effort~\cite{bommasani2022opportunities}. 
Prompt tuning methods~\cite{yao2023visual, zhang2023prompt, bahng2022exploring} can efficiently adapt VFMs towards dedicated scenarios by designing suitable prompts.
Feature adaptation methods~\cite{gao2023clip, zhang2021tip, svladapterbmvc2022, zhou2022extract, zhao2023vpd} can further pivot VFMs towards different tasks.
Direct tuning enables more flexible adaptation, especially in few-shot customization scenarios like DreamBooth~\cite{ruiz2023dreambooth}.
As we show in this paper, \method{} can be interpreted as an instance of this type of tuning, where Stable Diffusion~\cite{rombach2022high} plays the role of the foundation model for multiple image analysis tasks. 
Our models exhibit strong in-the-wild performance thanks to the foundational prior and efficient fine-tuning protocol~(\cf the teaser figure).

\subsection{Monocular Depth Estimation}
The pioneering work~\cite{eigen_depth_2014} introduced an end-to-end trainable network and showed that metric depth for a dataset can be recovered with a single sensor. 
Successive improvements have come from various fronts, including 
various parameterizations (ordinals, bins, planar maps, CRFs, \textit{etc.})~\cite{fu2018dorn, lee2019big, yuan2022newcrfs, patil2022p3depth, liu2023vadepth, Farooq_Bhat_2021, li2022binsformer, ning2023ait} and
switching CNN backbones to vision transformers~\cite{yang2021transformers, li2023depthformer, aich2021bidirectional,bhat2023zoedepth}. 
A class of works~\cite{yin2023metric3d,hu2024metric3d,guizilini2023towards_zero_depth} relies on privileged information fed alongside input, such as camera intrinsics. 
Estimating depth ``in the wild'' or ``out-of-distribution'' refers to methods with robustness across a wide range of possibly unfamiliar settings where no privileged information is available. 
MegaDepth~\cite{MDLi18_megadepth} and DiverseDepth~\cite{yin2020diversedepth} utilize extensive internet photo collections to train models that can adapt to unseen data, while MiDaS~\cite{Ranftl2020_midas} achieves generality by training on a mixture of multiple datasets.
To unify representations across datasets, MiDaS estimates affine-invariant depth -- up to unknown shift and scale.
The step from CNNs to vision transformers has further boosted performance, as evidenced by DPT~(MiDaS v3)~\cite{ranftl2021_dpt}.
Depth Anything~\cite{yang2024depthv1} took data scaling to the next level by relying on DINOv2~\cite{oquab2023dinov2} foundational model prior trained on 142M images with self-supervision and subsequent training with 62M pseudo-labels, 1M real depth annotations, and 0.5M synthetic ones.
Several methods~\cite{zhao2023vpd,duan2023diffusiondepth,saxena2023depthgen,zhao2023vpd} proposed using DDPMs and LDMs for depth estimation.
However, they either train models from scratch, use Stable Diffusion~\cite{rombach2022high} as a feature extractor, resort to custom latent spaces, operate in pixel space, or require extensive training.
Our previous work~\cite{ke2023repurposing} proposed fine-tuning a generative text-to-image LDM such as Stable Diffusion, trained with LAION-5B~\cite{schuhmann2022laion5b}, a dataset of 2.3B image-text pairs, towards affine-invariant depth using just 74K samples from the HyperSim~\cite{roberts2021hypersim} and Virtual KITTI~\cite{cabon2020virtualkitti2} synthetic datasets.
\method{} demonstrated impressive zero-shot generalization both in benchmarks and ``in the wild''.

Since uploading~\cite{ke2023repurposing},
Depth Anything V2~\cite{yang2024depthv2} confirmed our findings about the role of synthetic data in the task and retired real data from their pipeline, achieving impressive performance gains.
E2E-FT~\cite{garcia2024fine} and GenPercept~\cite{xu2024genpercept} performed end-to-end fine-tuning. 
The former also proposed a fix for a few-step DDIM scheduler inference regime. 
E2E-FT also demonstrated that end-to-end networks can score higher in zero-shot benchmarks than the similar generative model. 
Two more works kept the generative nature of the base model:
Lotus~\cite{he2024lotus} switched to predicting the target using exactly one step;
DepthFM~\cite{gui2024depthfm} adopted flow matching~\cite{lipman2022flow} at training. 
GeoWizard~\cite{fu2024geowizard} proposed to estimate the depth and surface normals jointly, although using privileged information about scene type. 
BetterDepth~\cite{zhang2024betterdepth} introduced a \method{}-based refiner for coarse depth inputs.
SteeredMarigold~\cite{gregorek2024steeredmarigold} used Marigold as a generative prior in order to perform depth densification.
Robustness under challenging conditions~\cite{tosi2024diffusion} was tackled through depth-conditioned generation of training data, an approach similar to DGInStyle~\cite{jia2024dginstyle}.
ChronoDepth~\cite{shao2024learning} and DepthCrafter~\cite{hu2024depthcrafter} address the temporal consistency of depth prediction in the video domain by performing \method{}-like fine-tuning of video diffusion models. 

We stick to the generative formulation of \method{}, incorporate the DDIM fix~\cite{garcia2024fine}, and recap the method~\cite{ke2023repurposing} (Sec.~\ref{sec:method}).

\subsection{Monocular Surface Normals Prediction}
Early monocular surface normals estimation methods often employed CNNs consisting of a feature extractor backbone followed by a prediction head~\cite{ladicky2014,wang2015deep,huang2019framenet,eigen2015}. 
Over time, various improvements have been proposed. 
Bae et al.~\cite{Bae2021aleatoric} suggested estimating aleatoric uncertainty and using uncertainty-guided sampling during training to enhance prediction quality for small structures and object boundaries. 
Omnidata v2~\cite{kar2022omnidata} introduced a transformer-based model trained on 12M images, applying sophisticated 3D data augmentation and enforcing cross-task consistency. 
DSINE~\cite{bae2024dsine} identified and incorporated inductive biases tailored for surface normals estimation. 
It leverages the per-pixel ray direction coupled with a ray-based activation function and learns the relative rotation between neighboring surface normals. 

Since uploading~\cite{ke2023repurposing}, several papers have repurposed diffusion models to surface normals estimation.
GeoWizard~\cite{fu2024geowizard} jointly estimates depth and surface normals, although it uses privileged information about scene type. 
Shortly after, we released \method{}-Normals \texttt{v0.1}, a preview model for monocular surface normals estimation trained similarly to \method{}-Depth, albeit just on HyperSim.
GenPercept~\cite{xu2024genpercept} treats the denoising U-Net as a deterministic backbone and employs one-step inference. 
StableNormal~\cite{ye2024stablenormal} aims to reduce the inherent stochasticity of diffusion models by using a two-stage coarse-to-fine strategy. 
A single-step surface normals estimator first produces an initial coarse estimate, followed by a refinement process that recovers finer geometric details, semantically guided by DINOv2~\cite{oquab2023dinov2} features. 
Lotus~\cite{he2024lotus} directly predicts annotations instead of noise and reformulates the multi-step diffusion into a single-step procedure. 
Garcia~\etal~\cite{garcia2024fine} proposed a single-step inference fix for \method{}'s DDIM scheduler to enhance inference speed.
Additionally, they reframe a single-step generative predictor into an end-to-end network based on the same architecture. 

In what follows, we continue developing the ideas behind \method{}-Normals and arrive at \texttt{v1.1}. 
We demonstrate that a careful curation of synthetic fine-tuning datasets together with the vanilla \method{} protocol without any other foundational models, multi-task aggregation, privileged information, or refinement stages achieves the best performance in most evaluation datasets, compared to the other recent diffusion-based surface normals estimation methods (Sec.~\ref{sec:method:normals}).

\subsection{Intrinsic Image Decomposition (IID)}
Intrinsic image decomposition aims to recover the intrinsic properties of objects in an image, including albedo (surface reflectance), shading, and Bidirectional Reflectance Distribution Function (BRDF) parameters, such as roughness and metallicity.
It was introduced by Horn~\cite{horn1970shape} and later studied by Barrow \etal~\cite{barrow1978recovering}.
The theory evolved from early prior-based approaches, such as the retinex theory, to modern deep learning-based methods.
First, deep learning approaches typically utilized a feed-forward convolutional network~\cite{li2020inverse, wang2021learning, li2022physically, luo2024intrinsicdiffusion} or a transformer~\cite{zhu2022irisformer} to predict pixel-level intrinsic decomposition channels from the input image. 
Careaga~\etal~\cite{careaga2024colorful} proposed a multi-step approach that first predicts initial albedo and grayscale shading maps using an off-the-shelf network~\cite{careaga2023intrinsic}, and then progressively refines them.
With the recent advent of vision foundation models, especially generative models, alternative solutions have shown success by utilizing StyleGAN~\cite{Karras_2019_CVPR, bhattad2024stylegan} or diffusion models~\cite{du2023generative, lee2023exploiting, kocsis2024intrinsic, zeng2024rgb}.
DMP~\cite{lee2023exploiting} learns a deterministic mapping between an input and the IID task (albedo and shading) through a low-rank adaptation of text-to-image models.
IID-Diffusion~\cite{kocsis2024intrinsic} uses a custom latent encoder and CLIP~\cite{radford2021learning} features to guide the fine-tuned Stable Diffusion~\cite{rombach2022high} on InteriorVerse~\cite{zhu2022interiorverse} decomposition into albedo and material properties: roughness and metallicity.
RGB$\leftrightarrow$X~\cite{zeng2024rgb} learns a bijection between input images and various modalities, including the ones used in IID-Diffusion.
Their model, based on the pretrained Stable Diffusion, has a similar architecture to \method{}, yet uses the text encoder to switch between pre-defined modalities.

We train two \method{}-IID models: one that predicts albedo, roughness, and metallicity, and another that estimates albedo, non-diffuse shading, and a residual diffuse component. 
We compare our models to IID-Diffusion~\cite{kocsis2024intrinsic}, RGB$\leftrightarrow$X~\cite{zeng2024rgb} and Careaga~\etal~\cite{careaga2024colorful} in their respective domains on the InteriorVerse~\cite{zhu2022interiorverse} and HyperSim~\cite{roberts2021hypersim} datasets.
Our simpler models achieve highly competitive performance in both quantitative and qualitative evaluations (Sec.~\ref{sec:method:iid}).

\subsection{High-Resolution Estimation}

Models fine-tuned from Stable Diffusion~\cite{rombach2022high} usually exhibit resolution bias towards the original resolution used to train the text-to-image LDM.
The same applies to all \method{} models: their processing resolution defaults to the recommended value of 768 inherited from the base model, which should correspond to the longest side of the input image.
As a result, large resolutions suffer a major loss of details during downsampling to the processing resolution and upsampling output to the original size (see the teaser figure).
This issue prompted us to investigate approaches to high-resolution inference with \method{}.
Although the study is centered around \method{}-Depth, the findings apply to other modalities.

BoostingDepth~\cite{miangoleh2021boosting} fuses local patches into a global canvas through an additional GAN network. 
PatchFusion~\cite{li2024patchfusion} introduces a tile-based framework that combines globally consistent coarse features with finer local features using a patch-wise fusion network. 
Moreover, it ensures consistency across patches during training without post-processing. 
PatchRefiner~\cite{li2024patchrefiner} performs high-resolution depth estimation by refining predictions with a pseudo-labeling strategy.
MultiDiffusion~\cite{bar2023multidiffusion} is an inference protocol for pre-trained diffusion models that generates high-resolution outputs by performing diffusion over overlapping tiles in an interleaved fashion.
While it has been used in generative tasks like semantic segmentation~\cite{jia2024dginstyle}, it has not been applied yet to enhance the resolution of image analysis tasks.
Depth Pro~\cite{appledepthpro} proposed concurrently a multi-scale vision transformer trained on tens of datasets, capable of predicting metric depth for HD images in a split second.

Our \method{}-HR inference strategy attains competitive or better performance compared to these other methods (Sec.~\ref{sec:method:hr}).

%% file: sec/3_method.tex
\section{Base Model: Marigold-Depth}
\label{sec:method}

\newcommand{\img}{\mathbf{x}}
\newcommand{\depth}{\mathbf{d}}
\newcommand{\latent}{\mathbf{z}}
\newcommand{\latentdepth}{\latent^{(\depth)}}
\newcommand{\latentimage}{\latent^{(\img)}}
\newcommand{\noise}{\bm{\epsilon}}
\newcommand{\denoiser}{\bm{\epsilon}_{\theta}}
\newcommand{\denoiserlong}{\denoiser(\latentdepth_t, \latentimage, t)}
\newcommand{\catinput}{\mathbf{z}}
\newcommand{\encoder}{\mathcal{E}}
\newcommand{\decoder}{\mathcal{D}}

In this section, we recap the details of our fine-tuning protocol in the context of monocular depth estimation~\cite{ke2023repurposing}.

\subsection{Generative Formulation}
\label{sec:preliminary}

We pose monocular depth estimation as a conditional denoising diffusion generation task and train \method{} to model the conditional distribution $D(\depth~|~\img)$ over depth $\depth \in \mathbb{R}^{W \times H}$, where the condition $\img \ {\in}\  \mathbb{R}^{W \times H \times 3}$ is an RGB image.

In the \textit{forward} process, which starts at $\depth_0 := \depth$ from the conditional distribution, Gaussian noise is gradually added at levels $t \in \{1,..., T\}$ 
to obtain noisy samples $\depth_t$ as
\begin{equation}
    \depth_t = \sqrt{\bar{\alpha}_t} \depth_0 + \sqrt{1 - \bar{\alpha}_t} \noise
\end{equation}
where $\noise \sim \mathcal{N}(0, I),
\bar{\alpha}_t := 
\prod_{s=1}^{t}{1 \!-\! \beta_s}$, 
and $\{\beta_1, \ldots, \beta_T\}$ is the variance schedule of a process with $T$ steps. 
In the \textit{reverse} process, the denoising model $\denoiser(\cdot)$ parameterized with learned parameters $\theta$ gradually removes noise from $\depth_t$ to obtain  $\depth_{t-1}$. 

At training time, parameters $\theta$ are updated by taking a data pair $(\img, \depth)$ from the training set, noising $\depth$ with sampled noise $\noise$ at a random timestep $t$, computing the noise estimate $\hat{\noise} = \denoiser(\depth_t, \img, t)$ and minimizing one of the denoising diffusion objective functions.
The canonical standard noise objective $\mathcal{L}$ is given as follows~\cite{ho2020denoising_ddpm}:
\begin{equation}
\mathcal{L} = \mathbb{E}_{\depth_0, \noise \sim \mathcal{N}(0,I),t \sim \mathcal{U}(T)} \left\| \noise - \hat{\noise} \right\|^2_2.
\label{eq:diffusion_objective}
\end{equation}
At inference time, $\depth := \depth_0$ 
is reconstructed starting from a normally-distributed variable $\depth_T$, by iteratively applying the learned denoiser $\bm{\epsilon}_\theta(\depth_t, \img, t)$. 

We consider the latent space formed in the bottleneck of a VAE, trained independently of the denoiser, to enable latent space compression and perceptual alignment with the data space.
To translate our formulation into the latent space, for a given depth map $\depth$, the corresponding latent code is given by the encoder $\encoder$: $\latentdepth = \encoder(\depth)$.
Given a depth latent code, a depth map can be recovered with the decoder $\decoder$: $\hat{\depth} = \decoder(\latentdepth)$.
The conditioning image $\img$ is also naturally translated into the latent space as $\latentimage = \encoder(\img)$.
The denoiser is henceforth trained in the latent space: $\denoiserlong$.
The adapted inference procedure involves one extra step -- the decoder $\decoder$ reconstructing the data $\hat{\depth}$ from the estimated clean latent $\latentdepth_0$: $\hat{\depth} = \decoder(\latentdepth_0)$.

\subsection{Network Architecture}
\label{sec:architecture}

\begin{figure}[t!]
    \centering
    \includegraphics[width=\linewidth]{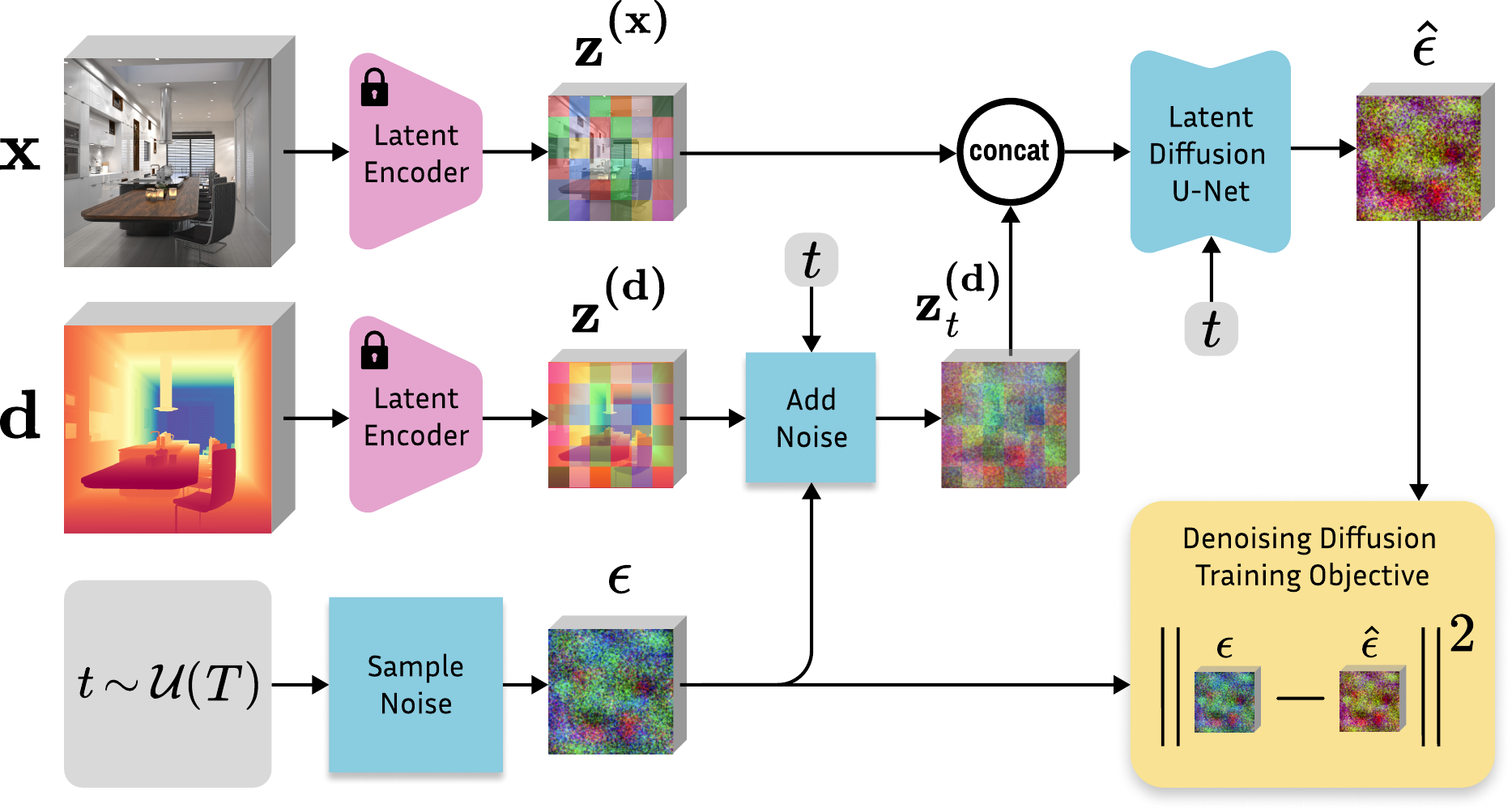}
    \caption{
    \textbf{Overview of the Marigold fine-tuning protocol.}
    Starting from a pretrained Stable Diffusion, we encode the image $\img$ and depth $\depth$ into the latent space using the original Stable Diffusion VAE.
    We fine-tune just the U-Net by optimizing the standard diffusion objective relative to the depth latent code.
    Image conditioning is achieved by concatenating the two latent codes before feeding them into the U-Net. 
    The first layer of the U-Net is modified to accept concatenated latent codes.
    See details in Sec.~\ref{sec:architecture} and Sec.~\ref{sec:finetuning}.
    }
    \label{fig:method-train}
\end{figure}

We base our model on a pretrained text-to-image LDM Stable Diffusion v2~\cite{rombach2022high}.
With minimal changes to the model, we turn it into a conditional depth map generator (Fig.~\ref{fig:method-train}).

\vspace{0.35em}
\noindent\textbf{%
Depth encoder and decoder.
}
We take the frozen VAE to encode \textit{both} the image and its corresponding depth map into a latent space for training our conditional denoiser. 
Given that the encoder, which is designed for 3-channel (RGB) inputs, receives a single-channel depth map, we replicate the depth map into three channels to simulate an RGB image.
At this point, the data range of the depth data plays a significant role in enabling affine-invariance. We discuss our normalization approach in Sec.~\ref{sec:finetuning}.
We verified that
without any modification of the VAE or the latent space structure, 
the depth map can be reconstructed from the encoded latent code with a negligible error, \ie, $\depth \approx \decoder(\encoder(\depth))$.
This is the first check to be performed when following the \method{} fine-tuning protocol for a new modality.
At inference time, the depth latent code is decoded once at the end of diffusion, and the average of three channels is taken as the predicted depth map.
Extending \method{} to another modality with a different number of channels may prompt allocating new latent space for each triplet of channels.

\vspace{0.35em}
\noindent\textbf{%
Adapted denoising U-Net.
}
To implement the conditioning of the latent denoiser $\denoiserlong$ on input image $\img$, we concatenate the image and depth latent codes into a single input $\catinput_t = \text{cat}(\latent^{(\depth)}_t, \latent^{(\img)})$ along the feature dimension. 
The input channels of the latent denoiser are then doubled to accommodate the expanded input $\catinput_t$. 
To prevent inflation of activations magnitude of the first layer and keep the pre-trained structure as faithfully as possible, we duplicate the weight tensor of the input layer and divide its values by two.
A similar conditioning mechanism, except for the zero weights initialization, was previously used in InstructPix2Pix~\cite{brooks2023instructpix2pix}.

\subsection{Fine-Tuning for Depth Estimation}

\label{sec:finetuning}

\noindent\textbf{%
Affine-invariant depth normalization.
}
For the ground truth depth maps $\depth$, we implement a linear normalization such that the depth primarily falls in the value range $[-1, 1]$,
to match the designed input value range of the VAE.
Such normalization serves two purposes.
First, it is the convention for working with the original Stable Diffusion VAE.
Second, it enforces a canonical affine-invariant depth representation independent of the data statistics -- any scene must be bounded by near and far planes with extreme depth values.
The normalization is achieved through an affine transformation computed as
\begin{equation}
\Tilde{\depth} = \left(\frac{\depth - \depth_{2}}{\depth_{98} - \depth_{2}} - 0.5\right) \times 2,
\end{equation}
where $\depth_{2}$ and $\depth_{98}$ correspond to the $2\%$ and $98\%$ percentiles of individual depth maps.
This normalization allows \method{} to focus on pure affine-invariant depth estimation.

\vspace{0.35em}
\noindent\textbf{%
Training on synthetic data.
}
Real depth datasets suffer from missing depth values caused by the physical constraints of the capture rig and the physical properties of the sensors. 
Specifically, the disparity between cameras and reflective surfaces diverting LiDAR laser beams are inevitable sources of ground truth noise and missing pixels~\cite{wagner2006gaussian,Huang2023nfl}.
In contrast to prior work that utilized diverse real datasets to achieve generalization~\cite{Ranftl2020_midas, eftekhar2021omnidata}, we train exclusively with synthetic depth datasets. 
As with the depth normalization rationale, this decision has two objective reasons.
First, synthetic depth is inherently dense and complete, meaning that every pixel has a valid ground truth depth value, allowing us to feed such data into the VAE, which can not handle data with invalid pixels.
Second, synthetic depth is the cleanest possible form of depth, which is guaranteed by the rendering pipeline. 
It provides the cleanest examples and reduces noise in gradient updates during the short fine-tuning protocol.
Thus, the remaining concern is the sufficient diversity or domain gaps between synthetic and real data, which sometimes limits generalization ability.
As demonstrated in our experiments across modalities, our choice of synthetic datasets leads to impressive zero-shot transfer. 

\begin{figure}[t!]
    \centering
    \includegraphics[width=\linewidth]{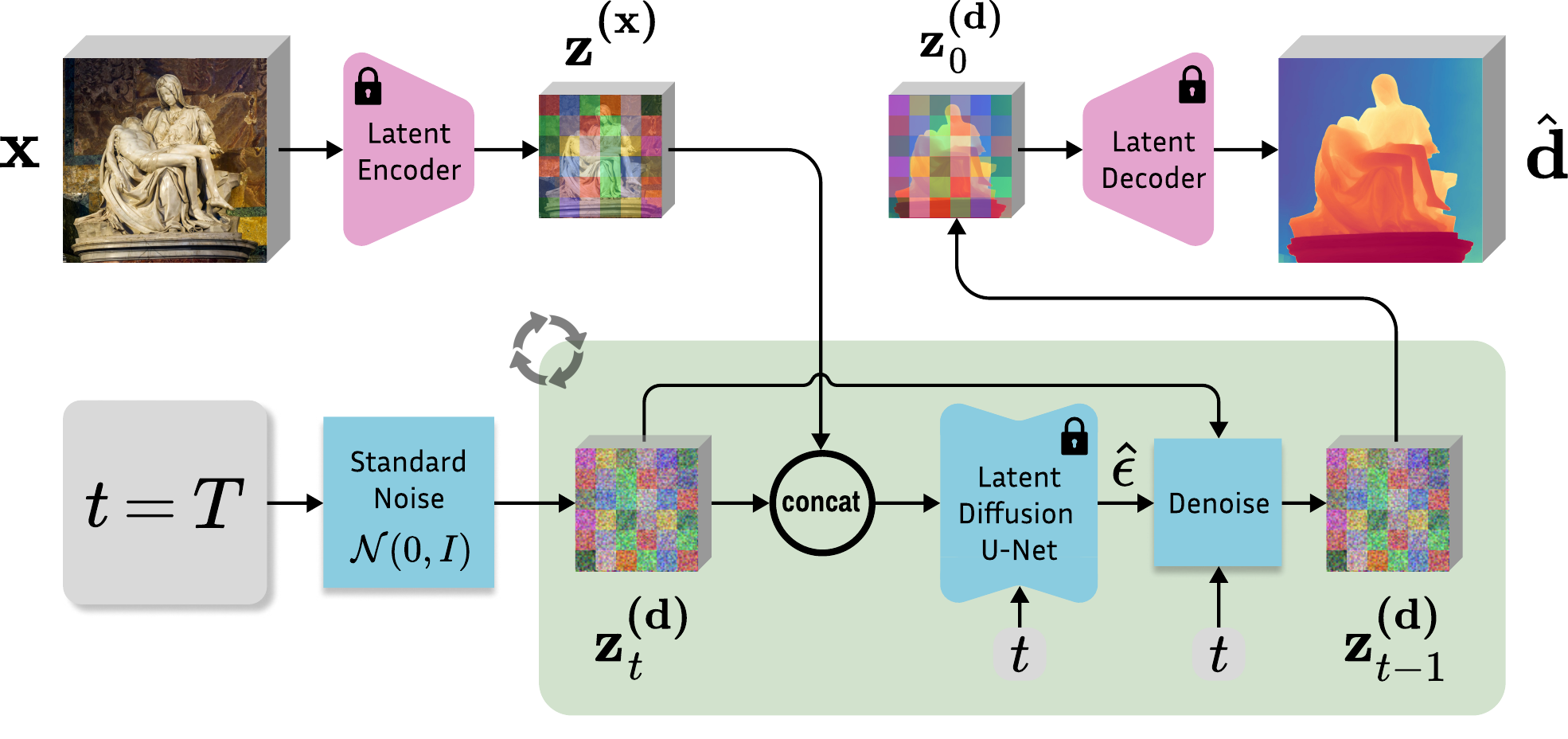}
    \caption{
    \textbf{Overview of the Marigold inference scheme.}
    Given an image $\img$, we encode it with the original Stable Diffusion VAE into the latent code $\latentimage$, and concatenate with the depth latent $\latentdepth_t$ before giving it to the modified fine-tuned U-Net on every denoising iteration.
    After executing the schedule of $T$ steps, the resulting depth latent $\latentdepth_0$ is decoded into an image whose 3 channels are averaged to get the final estimation $\hat\depth$. 
    See Sec.~\ref{sec:inference} for details.
    }
    \label{fig:method-inference}
\end{figure}

\subsection{Inference}
\label{sec:inference}

\noindent\textbf{%
Latent diffusion denoising.
}
The overall inference pipeline is presented in Fig.~\ref{fig:method-inference}.
We encode the input image into the latent space, initialize depth latent as standard Gaussian noise, and progressively denoise it with the same schedule as during fine-tuning.
We use DDIM~\cite{song2020denoising_ddim} to perform non-Markovian sampling with re-spaced steps for accelerated inference.
The final depth map is decoded from the latent code using the VAE decoder and postprocessed by averaging channels.

\newcommand{\pred}{\mathbf{\hat{d}}}
\newcommand{\translated}{\mathbf{\hat{d^{\prime}}}}
\newcommand{\merged}{\mathbf{m}}

\vspace{0.35em}
\noindent\textbf{%
Test-time ensembling.
}
The stochastic nature of the inference pipeline leads to varying predictions depending on the initialization noise in $\latentdepth_T$.
Capitalizing on that, we propose the following test-time ensembling scheme, capable of combining multiple inference passes over the same input.
For each input sample, we can run inference $N$ times.
To aggregate these affine-invariant depth predictions $\{ \pred_1, \ldots, \pred_N\}$, we jointly estimate the corresponding scale $\hat{s}_i$ and shift $\hat{t}_i$, relative to some canonical scale and range, in an iterative manner.
The proposed objective minimizes the distances between each pair of scaled and shifted predictions $(\translated_i, \translated_j)$, where $\translated = \pred \times \hat{s}+ \hat{t}$. 
In each optimization step, we calculate the merged depth map $\merged$ by the taking pixel-wise median $\merged(x, y) = \text{median}(\translated_1(x, y), \ldots, \translated_N(x, y))$.
An extra regularization term $\mathcal{R} = |\! \min (\merged)| + |1 - \max (\merged)|$, is added to prevent collapse to the trivial solution and enforce the unit scale of $\merged$.
Thus, the objective function can be written as follows:
\begin{equation}
      \min_{\substack{s_1,\ldots,s_N \\ t_1, \ldots, t_N}} \Bigg( \sqrt{\frac{1}{b} \sum_{i=1}^{N-1} \sum_{j=i+1}^{N} \| \translated_i - \translated_j \|_2^2} + \lambda \mathcal{R} \Bigg)   
      \label{eq:align}
\end{equation}
where the binominal coefficient $b = \binom{N}{2}$ represents the number of possible combinations of image pairs from $N$ images.
After the iterative optimization for spatial alignment, the merged depth $\merged$ is taken as our ensembled prediction.
Note that this ensembling step requires no ground truth for aligning independent predictions.
This scheme enables a flexible trade-off between computation efficiency and prediction quality by choosing $N$ accordingly.

\subsection{Implementation} \label{sec:implementation}

We use Stable Diffusion v2~\cite{rombach2022high} as base LDM, following the original pre-training setup with a \mbox{$v$-objective}~\cite{salimans2022progressive}. 
We disable text conditioning and perform steps outlined in Sec.~\ref{sec:architecture}.
During training, we apply the DDPM noise scheduler~\cite{ho2020denoising_ddpm} with 1000 diffusion steps. 
At inference time, we apply DDIM scheduler~\cite{song2020denoising_ddim} and sample between 1 and 50 steps. 
To approximate the mode of conditional distribution and increase quality, we ensemble 10 predictions from different initial noise.
Training takes 18K iterations using a batch size of 32. 
To fit one GPU, we use a real batch size of 2 and accumulate gradients 16 times. 
We use the Adam optimizer with a $3\cdot 10^{-5}$ learning rate. 
Additionally, we apply random horizontal flipping augmentation to the training data. 
Training our method to convergence takes approximately 2.5 days on a single Nvidia RTX 4090 GPU card.
Unlike the $10 \times 50$ zero-shot evaluation protocol, inference with one ensemble member and one diffusion step $1 \times 1$ produces sufficiently good results fast, often sharper than the ensembled prediction. 
Coupled with model weight quantization and smaller compatible VAEs, such as TAESD~\cite{taesd}, the $1\times 1$ prediction takes less than 100ms on most hardware.

\begin{table*}[t!]
    \centering
    \caption{
        \textbf{Quantitative comparison}
        of \method{}-Depth with SOTA affine-invariant depth estimators on several zero-shot benchmarks$^1$.
        In the 1$^{\mathrm{st}}$ section, we list methods citing our approach~\cite{ke2023repurposing} as well as methods that require more than 2M samples for fine-tuning (denoted by \nocompete{gray}).
        We compare methods from the 2$^{\mathrm{nd}}$ section with flavors of \method{}-Depth \texttt{v1.0} (ours, CVPR'2024) from the 3$^{\mathrm{rd}}$ section and \method{}-Depth \texttt{v1.1} (ours, this paper) in the 4$^{\mathrm{th}}$ section.
        Legend:
        All metrics are presented in percentage terms; 
        bold numbers are the best, underscored second best;
        NFEs is the number of function evaluations required to obtain the prediction -- $ensemble \times steps$ for diffusion models and $1$ for end-to-end networks.
        \method{} outperforms other methods in this low-data regime on indoor and outdoor scenes without access to real depth samples.
    }
    \resizebox{\linewidth}{!}{
	\input{tbl/1_qualitative}
	\label{table:zeroshot_test}
    }
    \\
    \begin{minipage}{0.98\linewidth}
        \scriptsize
        \vspace{0.4em}
        \begin{itemize}
        \item[$^1$]
            Metrics in the 1$^\mathrm{st}$ section are sourced from the respective papers.
            Metrics in the 2$^\mathrm{nd}$ section are sourced from Metric3D~\cite{yin2023metric3d}, except the ScanNet benchmark. 
            For ScanNet, Metric3D used a different random split that is not publicly accessible. 
            Therefore, we re-ran baseline methods on our split. 
            We additionally took numbers from Metric3D for HDN~\cite{zhang2022_hdn} on ScanNet benchmark due to unavailable source code. 
        \item[$^2$]
            Privileged information used by methods: 
            Metric3D and Metric3D v2 require camera intrinsics; 
            GeoWizard requires choosing between indoor and outdoor regimes.
        \item[$^3$]
            These Marigold variants are evaluated using the trailing timestamps setting of the DDIM scheduler.
        \end{itemize}
    \end{minipage}
\end{table*}

 \begin{table}[t]
    \caption{\textbf{Inference time} of \method{}-Depth and other methods on a $768 \times 768$ image using an RTX 3090 GPU.}
    \label{tab:depth_speed}
    \resizebox{\linewidth}{!}{
        \input{tbl/8_depth_speed_v2}
	\label{table:speed_test}
    }
 \end{table}

\subsection{Evaluation} 
\label{sec:evaluation}

\noindent\textbf{%
Training datasets.
} 
We employ two synthetic datasets covering both indoor and outdoor scenes.
\textbf{HyperSim}~\cite{roberts2021hypersim} is a photorealistic dataset with 461 indoor scenes. 
We use the official split for training, with around 54K samples from 365 scenes, filtering out incomplete samples.
RGB images and depth maps are resized to $480 \times 640$ resolution.
Depth is normalized with the dataset statistics.
Additionally, we transform distances relative to the focal point into depth values relative to the focal plane. 
The second dataset, \textbf{Virtual KITTI}~\cite{cabon2020virtualkitti2}, is a synthetic street-scene dataset featuring 5 scenes with diverse weather and camera perspectives. 
We crop the images to the KITTI resolution~\cite{Geiger2012CVPR} and set the far plane to 80 meters.
New in \method{}-Depth \texttt{v1.1}: (1) the training data is augmented with flipping, blurring, and color jitter; (2) DDIM timesteps are set to ``trailing'' and zero SNR is enabled~\cite{garcia2024fine} before fine-tuning.

\vspace{0.35em}
\noindent\textbf{%
Evaluation datasets.
}
We evaluate \method{}-Depth on 5 real datasets not seen during training. 
\textbf{NYUv2}~\cite{SilbermanECCV12nyu} and \textbf{ScanNet}~\cite{dai2017scannet} are both indoor scene datasets captured with an RGB-D Kinect sensor. 
For NYUv2, we utilize the designated test split, comprising 654 images. 
In the case of the ScanNet dataset, we randomly sampled 800 images from the 312 official validation scenes for testing.  
\textbf{KITTI}~\cite{Geiger2012CVPR} is a street-scene dataset with sparse metric depth captured by a LiDAR sensor. 
We employ the Eigen test split~\cite{eigen_depth_2014} made of 652 images.
\textbf{ETH3D}~\cite{schops2017multiEth3d} and \textbf{DIODE}~\cite{diode_dataset} are two high-resolution datasets, both featuring depth maps derived from LiDAR sensor measurements. 
For ETH3D, we incorporate all 454 samples with available ground truth depth maps. 
For DIODE, we use the entire validation split, which encompasses 325 indoor samples and 446 outdoor samples.

\vspace{0.35em}
\noindent\textbf{%
Evaluation protocol.
} 
Following the protocol of affine-invariant depth evaluation~\cite{Ranftl2020_midas}, we first align the estimated merged prediction $\merged$ to the ground truth $\depth$ with the least squares fitting. 
This step gives us the metric depth map $\mathbf{a} = \merged \times s + t$ in the same units as the ground truth.
Next, we apply two metrics~\cite{yin2023metric3d, Ranftl2020_midas, ranftl2021_dpt, Wei2021CVPR_leres} for assessing quality of depth estimation.
The first is Absolute Mean Relative Error (AbsRel~↓), calculated as: $\frac{1}{M} \sum_{i=1}^M {|\mathbf{a}_i - \depth_i|} / {\depth_i}$, where $M$ is the total number of pixels. 
The second, Threshold Accuracy ($\delta1$~↑), measures the proportion of pixels satisfying $\max({\mathbf{a}_i}/{\depth_i}, {\depth_i}/{\mathbf{a}_i}) < 1.25$.

\vspace{0.35em}
\noindent\textbf{%
Comparison with other methods.
}
We compare \method{}-Depth to 5 zero-shot baselines in Tab.~\ref{table:zeroshot_test}.
We filtered baselines based on the affordability of the training protocol (at most 2M training samples) and temporal relevance. 
\method{}-Depth, built upon the Stable Diffusion prior and a small set of synthetic samples, outperforms prior work and remains competitive with methods that rely on larger training sets. 

Although trained exclusively on synthetic depth datasets, the model generalizes well to a wide range of real-world scenes.
For visual assessment, we present a qualitative comparison and 3D visualizations of surface normals reconstructed from depth in Figs.~\ref{fig:depth_comparison} and~\ref{fig:normal_comparison3d}. 
\method{} accurately captures both the overall scene layout, such as the spatial relationships between walls and furniture, and fine-grained details, as indicated by the arrows. 
In particular, the reconstruction of flat surfaces, especially walls, is significantly improved. 
Additionally, we report an inference speed comparison in Tab.~\ref{table:speed_test}.

\begin{figure*}[t]
    \centering
    \setlength\tabcolsep{0.5pt}
        \include{fig/4_qualitative_depth}
    \caption{
        \textbf{Qualitative comparison}
        of monocular depth estimation methods across different datasets. 
        \method{} excels at capturing thin structures~(\eg, chair legs) and preserving overall layout of the scene~(\eg, walls in ETH3D and chairs in DIODE). 
    }
    \label{fig:depth_comparison}
\end{figure*}

\begin{figure}[t!]
    \centering
    \includegraphics[width=0.99\linewidth]{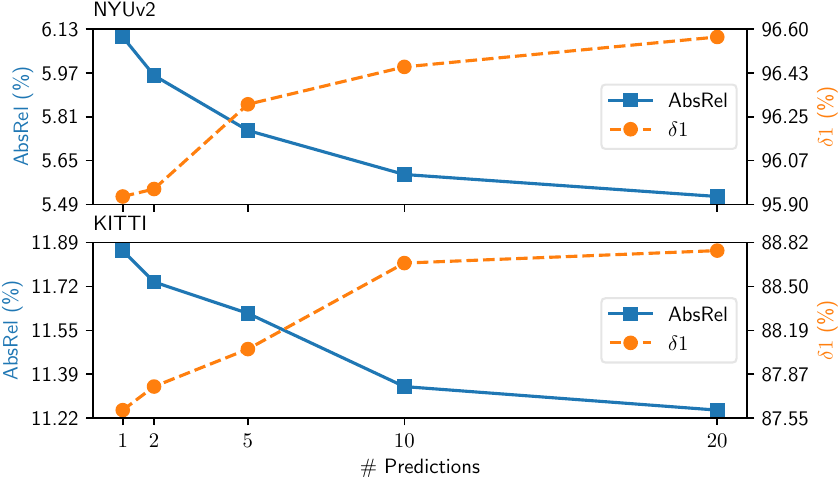}
    \caption{
        \textbf{Ablation of ensemble size.} 
        We observe a monotonic improvement with the growth of ensemble size. This improvement starts to diminish after 10 predictions per sample. 
    }
   \label{fig:merge_pred}
\end{figure}

\begin{figure}[t!]
    \centering
    \setlength\tabcolsep{0.5pt}

\include{fig/5_qualitative_normal}

    \caption{
        \textbf{Qualitative comparison (unprojected from depth, colored as normals)}
        of monocular depth estimation methods across different datasets. 
        \method{}-Depth stands out for its superior reconstruction of flat surfaces and detailed structures. 
    }
    \label{fig:normal_comparison3d}
\end{figure}

\begin{figure}[t!]
    \centering
    \includegraphics[width=0.99\linewidth]{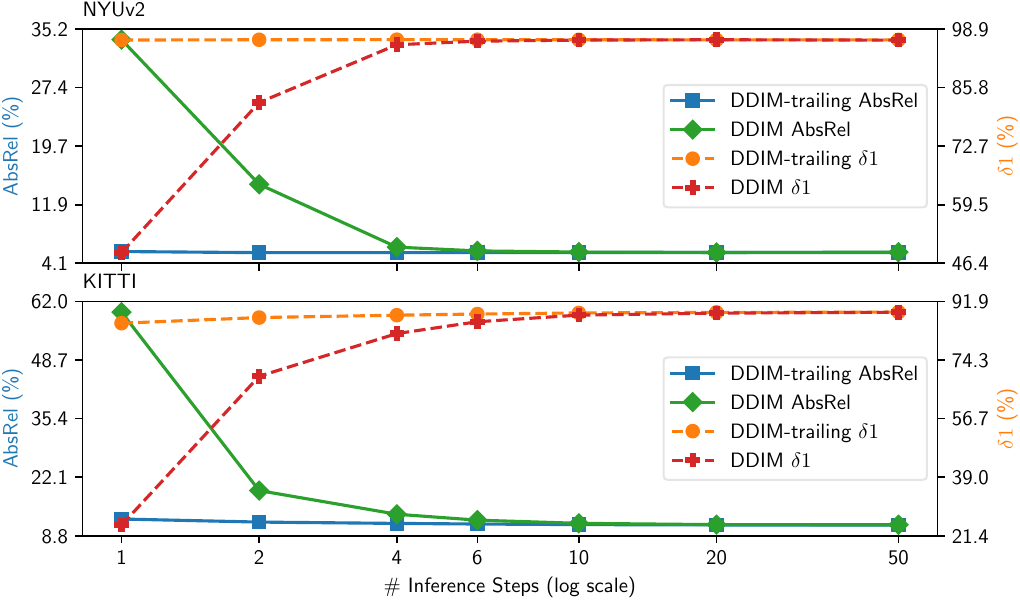}
    \caption{
        \textbf{Ablation of denoising steps.}
        One denoising step is sufficient with DDIM-trailing~\cite{garcia2024fine} (default in \texttt{v1.1}).
        DDIM~\cite{ke2023repurposing} (default in \texttt{v1.0}) requires at least 4-10 steps.
    }
   \label{fig:inference_denoise_step}
\end{figure}

\begin{table}[t]
    \centering
    \caption{\textbf{Training datasets:} 
        HyperSim~\cite{roberts2021hypersim} alone contributes the most;
        Virtual KITTI~\cite{cabon2020virtualkitti2} improves outdoor results.
    }
    \resizebox{\linewidth}{!}{
        \input{tbl/3_training_dataset}
    }
    \label{tbl:train_domain}
\end{table}

\subsection{Ablation Studies}
\label{sec:ablation}

We select two zero-shot validation sets for ablation studies: the official training split of NYUv2~\cite{SilbermanECCV12nyu}, consisting of 785 samples, and a randomly selected subset of 800 images from the KITTI Eigen~\cite{eigen_depth_2014} training split.

\vspace{0.35em}
\noindent\textbf{%
Training data domain.
}
To better understand the impact of the synthetic data on model generalization, we conduct an ablation study using the two datasets employed during training.
The results, presented in Tab.~\ref{tbl:train_domain}, show that even fine-tuning on a single synthetic dataset enables the pretrained LDM to adapt reasonably well to monocular depth estimation.
However, more diverse and photorealistic data yield stronger performance across both indoor and outdoor scenes. 
Notably, incorporating training data from a different domain enhances performance not only on that domain but also on the original one.

\vspace{0.35em}
\noindent\textbf{%
Test-time ensembling.
}
We test the effectiveness of the proposed test-time ensembling scheme by varying the number of predictions.
As shown in Fig.~\ref{fig:merge_pred}, a single prediction already yields reasonably good results. 
Ensembling 10 predictions can reduce the absolute relative error on NYUv2 by ${\sim}8\%$, and ensembling 20 predictions brings an improvement of ${\sim}9.5\%$.

\vspace{0.35em}
\noindent\textbf{%
Number of denoising steps.
}
Like the base model, \method{} is configured to use a 1000-step DDPM schedule during training.
In version \texttt{1.0}, inference used DDIM with leading timesteps, requiring 4 to 10 function evaluations (NFE) to reach peak performance. 
To report best-case results, we used 50 steps in the initial evaluations. 
Later, Garcia~\textit{et al.}~\cite{garcia2024fine} proposed switching to trailing timesteps for inference with DDIM. 
This change significantly improved efficiency of \method{}, with performance saturating at just 1 DDIM step (NFE=1). 
This is the default in all \method{} \texttt{v1.1} models. 
Table~\ref{table:zeroshot_test} reports results across model versions, NFE, ensemble sizes, alternative VAEs, and FP16 quantized weights.
The effect of varying denoising steps in DDIM scheduler~\cite{song2020denoising_ddim} is shown in Fig.~\ref{fig:inference_denoise_step}.

\input{sec/3_method_normals}

\input{sec/3_method_albedo}

\input{sec/3_method_lcm}

\input{sec/3_method_hr}

%% file: tbl/1_qualitative.tex
\begin{tabular}{
@{}l @{} %
r@{\hspace{0.5em}} %
c@{\hspace{2em}} %
c@{\hspace{0.25em}} %
c@{\hspace{0.25em}} %
c@{\hspace{0.25em}} %
c @{}p{2.0em}@{} %
c@{\hspace{0.5em}}c @{}p{2.0em}@{} 
c@{\hspace{0.5em}}c @{}p{2.0em}@{} 
c@{\hspace{0.5em}}c @{}p{2.0em}@{} 
c@{\hspace{0.5em}}c @{}p{2.0em}@{} 
c@{\hspace{0.5em}}c 
}

\toprule

\multirow{2}{*}{Method} &
\multirow{2}{*}{NFEs} &
\multirow{2}{*}{Prior} &
\multicolumn{4}{c}{Data} & &
\multicolumn{2}{c}{NYUv2} & &
\multicolumn{2}{c}{KITTI} & &
\multicolumn{2}{c}{ETH3D} & &
\multicolumn{2}{c}{ScanNet} & &
\multicolumn{2}{c}{DIODE} 
\\

& 
& &
Prior & 
Generated &
Real & 
Synthetic & &
AbsRel ↓ & 
$\delta$1 ↑ & &
AbsRel ↓ & 
$\delta$1 ↑ & &
AbsRel ↓ & 
$\delta$1 ↑ & &
AbsRel ↓ & 
$\delta$1 ↑ & &
AbsRel ↓ & 
$\delta$1 ↑ 
\\

\midrule

\nocompete{Omnidata}~\cite{eftekhar2021omnidata} & 
\nocompete{1} & 
\nocompete{\scriptsize ImageNet} &
\nocompete{14M} &
\nocompete{---} &
\nocompete{12M} & 
\nocompete{310K} & &
\nocompete{7.4} & 
\nocompete{94.5} & &
\nocompete{14.9} & 
\nocompete{83.5} & &
\nocompete{16.6} & 
\nocompete{77.8} & &
\nocompete{7.5} & 
\nocompete{93.6} & &
\nocompete{33.9} & 
\nocompete{74.2} 
\\

\nocompete{Metric3D}~\cite{yin2023metric3d}\nocompete{$^2$} &
\nocompete{1} & 
\nocompete{\scriptsize ImageNet} &
\nocompete{14M} &
\nocompete{---} &
\nocompete{8M} & 
\nocompete{---} & &
\nocompete{5.8} &
\nocompete{96.3} & &
\nocompete{5.3} &
\nocompete{96.5} & &
\nocompete{6.4} &  %
\nocompete{96.5} & &
\nocompete{7.4} &  %
\nocompete{94.2} & &
\nocompete{21.1} &  %
\nocompete{82.5} 
\\

\nocompete{Metric3D v2}~\cite{hu2024metric3d}\nocompete{$^2$} &
\nocompete{1} & 
\nocompete{DINOv2} &
\nocompete{142M} &
\nocompete{---} &
\nocompete{16M} & 
\nocompete{91K} & &  %
\nocompete{4.3} &
\nocompete{98.1} & &
\nocompete{4.4} &
\nocompete{98.2} & &
\nocompete{4.2} &  %
\nocompete{98.3} & &
\nocompete{\sout{2.2}} &  %
\nocompete{\sout{99.4}} & &
\nocompete{13.6} &  %
\nocompete{89.5} 
\\

\nocompete{DepthAnything}~\cite{yang2024depthv1} &
\nocompete{1} & 
\nocompete{DINOv2} &  %
\nocompete{142M} &
\nocompete{62M} &
\nocompete{1M} &
\nocompete{524K} & &
\nocompete{4.3} &  %
\nocompete{98.1} & &
\nocompete{7.6} &  %
\nocompete{94.7} & &
\nocompete{12.7} &  %
\nocompete{88.2} & &
\nocompete{4.2} &  %
\nocompete{98.0} & &
\nocompete{27.7} &  %
\nocompete{75.9} 
\\

\nocompete{DepthAnything v2}~\cite{yang2024depthv2} &
\nocompete{1} & 
\nocompete{DINOv2} &  %
\nocompete{142M} &
\nocompete{62M} & 
\nocompete{---} &
\nocompete{595K} & &
\nocompete{4.4} &  %
\nocompete{97.9} & &
\nocompete{7.5} &  %
\nocompete{94.8} & &
\nocompete{13.2} &  %
\nocompete{86.2} & &
\nocompete{---} &  %
\nocompete{---} & &
\nocompete{6.5} &  %
\nocompete{95.4} 
\\

\nocompete{DepthFM}~\cite{gui2024depthfm} &
\nocompete{$? \times 1$} &
\nocompete{SD v2.1} &
\nocompete{2.3B} &
\nocompete{---} &
\nocompete{---} & 
\nocompete{63K} & & 
\nocompete{6.5} & 
\nocompete{95.6} & & 
\nocompete{8.3} & 
\nocompete{93.4} & &
\nocompete{---} & 
\nocompete{---} & &
\nocompete{---} & 
\nocompete{---} & &
\nocompete{22.5} & 
\nocompete{80.0} 
\\

\nocompete{GeoWizard}~\cite{fu2024geowizard}\nocompete{$^2$} & 
\nocompete{$10 \times 50$} &
\nocompete{SD v2.0} &
\nocompete{2.3B} &
\nocompete{---} &
\nocompete{51K} &  %
\nocompete{227K} & &  %
\nocompete{5.2} & 
\nocompete{96.6} & & 
\nocompete{9.7} & 
\nocompete{92.1} & &
\nocompete{6.4} & 
\nocompete{96.1} & &
\nocompete{6.1} & 
\nocompete{95.3} & &
\nocompete{29.7} & 
\nocompete{79.2} 
\\

\nocompete{GenPercept}~\cite{xu2024genpercept} & 
\nocompete{1} &  %
\nocompete{SD v2.1} &
\nocompete{2.3B} &
\nocompete{---} &
\nocompete{---} & 
\nocompete{74K} & & 
\nocompete{5.6} & 
\nocompete{96.0} & & 
\nocompete{9.9} & 
\nocompete{90.4} & &
\nocompete{6.2} & 
\nocompete{95.8} & &
\nocompete{---} & 
\nocompete{---} & &
\nocompete{35.7} & 
\nocompete{75.6} 
\\

\nocompete{Lotus-G}~\cite{he2024lotus} & 
\nocompete{$1 \times 1$} &  %
\nocompete{SD v2.0} &
\nocompete{2.3B} &
\nocompete{---} &
\nocompete{---} & 
\nocompete{59K} & & 
\nocompete{5.4} & 
\nocompete{96.6} & & 
\nocompete{11.3} & 
\nocompete{87.7} & &
\nocompete{6.2} & 
\nocompete{96.1} & &
\nocompete{6.0} & 
\nocompete{96.0} & &
\nocompete{---} & 
\nocompete{---} 
\\

\nocompete{Lotus-D}~\cite{he2024lotus} & 
\nocompete{1} &  %
\nocompete{SD v2.0} &
\nocompete{2.3B} &
\nocompete{---} &
\nocompete{---} & 
\nocompete{59K} & & 
\nocompete{5.3} & 
\nocompete{96.7} & & 
\nocompete{9.3} & 
\nocompete{92.8} & &
\nocompete{6.8} & 
\nocompete{95.3} & &
\nocompete{6.0} & 
\nocompete{96.3} & &
\nocompete{---} & 
\nocompete{---} 
\\

\nocompete{E2E-FT~\cite{garcia2024fine}} &
\nocompete{1} &  %
\nocompete{Marigold} &
\nocompete{2.3B} &
\nocompete{---} &
\nocompete{---} & 
\nocompete{74K} & & 
\nocompete{5.2} & 
\nocompete{96.6} & & 
\nocompete{9.6} & 
\nocompete{91.9} & &
\nocompete{6.2} & 
\nocompete{95.9} & &
\nocompete{5.8} & 
\nocompete{96.2} & &
\nocompete{30.2} & 
\nocompete{77.9} 
\\

\nocompete{E2E-FT~\cite{garcia2024fine}} &
\nocompete{1} &  %
\nocompete{SD v2.0} &
\nocompete{2.3B} &
\nocompete{---} &
\nocompete{---} & 
\nocompete{74K} & & 
\nocompete{5.4} & 
\nocompete{96.5} & & 
\nocompete{9.6} & 
\nocompete{92.1} & &
\nocompete{6.4} & 
\nocompete{95.9} & &
\nocompete{5.8} & 
\nocompete{96.5} & &
\nocompete{30.3} & 
\nocompete{77.6} 
\\

\midrule

DiverseDepth~\cite{yin2020diversedepth} & 
1 & 
\scriptsize{ImageNet} &
1M &
--- &
320K & 
--- & & 
11.7 & 
87.5 & & 
19.0 & 
70.4 & &
22.8 & 
69.4 & &
10.9 & 
88.2 & &
37.6 & 
63.1 
\\

MiDaS~\cite{Ranftl2020_midas} & 
1 & 
\scriptsize{ImageNet} &
1M &
--- &
2M & 
--- & &
11.1 & 
88.5 & &
23.6 & 
63.0 & &
18.4 & 
75.2 & &
12.1 & 
84.6 & &
33.2 & 
71.5 
\\

LeReS~\cite{Wei2021CVPR_leres} & 
1 & 
\scriptsize{ImageNet} &
1M &
--- &
300K & 
54K & &
9.0 & 
91.6 & &
14.9 & 
78.4 & &
17.1 & 
77.7 & &
9.1 & 
91.7 & &
27.1 & 
76.6 
\\

HDN~\cite{zhang2022_hdn} & 
1 & 
\scriptsize{ImageNet} &
14M &
--- &
300K & 
--- & &
{6.9} & %
{94.8} & &
11.5 & 
86.7 & &
12.1 & 
83.3 & &
8.0 & 
{93.9} & &
\underline{24.6} & 
\underline{78.0}
\\

DPT~\cite{ranftl2021_dpt} & 
1 & 
\scriptsize{ImageNet} &
14M &
--- &
1.2M & 
188K & &
9.8 & 
90.3 & &
\underline{10.0} & 
{90.1} & &
{7.8} & 
{94.6} & &
8.2 & 
93.4 & &
\textbf{18.2} & 
75.8 
\\

\midrule 

Marigold \texttt{v1.0}\nocompete{$^3$} w/ TAESD~\cite{taesd} & 
$1 \times 1$ &
SD v2.0&
2.3B &
--- &
--- &
74K & &
5.9 &
96.0 & &
12.2 &
86.2 & &
7.5 &
94.4 & &
6.7 &
95.0 & &
31.7 &
75.6
\\

Marigold \texttt{v1.0} LCM & 
$10 \times 1$ &
SD v2.0&
2.3B &
--- &
--- & 
74K & &
5.8 & %
{96.1} & &
{10.1} & 
\underline{90.9} & &
\underline{6.6} & 
\underline{95.8} & &
6.6 & 
95.0 & &
30.5 & 
77.2 
\\

Marigold \texttt{v1.0}\nocompete{$^3$} & 
$1 \times 1$ &
SD v2.0&
2.3B &
--- &
--- &
74K & &
\underline{5.7} &
\underline{96.2} & &
11.0 &
89.1 & &
6.9 &
95.5 & &
6.6 &
95.2 & &
31.2 &
76.6 
\\

Marigold \texttt{v1.0}\nocompete{$^3$} & 
$10 \times 1$ &
SD v2.0&
2.3B &
--- &
--- & 
74K & &
\underline{5.7} &
\underline{96.2} & &
10.9 &
89.2 & &
6.8 &
95.6 & &
6.5 &
95.3 & &
31.0 &
76.7  
\\

Marigold \texttt{v1.0} & 
$10 \times 50$ &
SD v2.0&
2.3B &
--- &
--- & 
74K & &
\textbf{5.5} & %
\textbf{96.4} & & %
\textbf{9.9} & 
\textbf{91.6} & &
\textbf{6.5} & 
\textbf{96.0} & &
6.4 & 
95.1 & &
30.8 & 
77.3
\\

\midrule 
Marigold \texttt{v1.1}\nocompete{$^3$} w/ TAESD~\cite{taesd} (fp16) & 
$1 \times 1$ &
SD v2.0 &
2.3B &
--- &
--- &
74K & &
6.1 &
95.8 & &
12.4 & 
85.0 & &
7.6 &
94.3 & &
6.8 & 
95.1 & &
31.1 & 
75.9
\\

Marigold \texttt{v1.1}\nocompete{$^3$} (fp16) & 
$1 \times 1$ &
SD v2.0 &
2.3B &
--- &
--- &
74K & &
5.8 &
96.1 & &
11.0 & 
88.8 & &
7.0 &
95.5 & &
6.6 &
95.3 & &
30.4 & 
77.3
\\

Marigold \texttt{v1.1}\nocompete{$^3$} & 
$1 \times 1$ &
SD v2.0 &
2.3B &
--- &
--- &
74K & &
5.9 &
96.1 & &
11.0 & 
88.8 & &
7.0 &
95.5 & &
6.6 &
95.3 &&
30.4 & 
77.3 
\\

Marigold \texttt{v1.1}\nocompete{$^3$} & 
$10 \times 1$ &
SD v2.0 &
2.3B &
--- &
--- &
74K & &
5.8 &
96.1 & &
10.9 & 
89.0 & &
6.9 &
95.7 & &
6.5 &
95.4 & &
30.3 &
77.3 
\\

Marigold \texttt{v1.1}\nocompete{$^3$} & 
$1 \times 4$ &
SD v2.0&
2.3B &
--- &
--- &
74K & &
\underline{5.7} &
\underline{96.2} & &
10.8 &
89.6 & &
7.2 &
95.3 & &
\underline{6.0} &
\underline{96.0} & &
30.1 &
77.9
\\

Marigold \texttt{v1.1}\nocompete{$^3$} & 
$10 \times 4$ &
SD v2.0&
2.3B &
--- &
--- &
74K & &
\textbf{5.5} &
\textbf{96.4} & &
10.5 &
90.2 & &
6.9 &
95.7 & &
\textbf{5.8} &
\textbf{96.3} & &
29.8 &
\textbf{78.2}
\\

\bottomrule

\end{tabular}

%% file: tbl/8_depth_speed_v2.tex
\begin{tabular}{l@{\hspace{10em}}c}
\toprule
Method & Time (sec) \\
\midrule
DPT        &  0.158  \\
DepthAnything v2       & 0.289 \\
Metric3D v2 & 0.386 \\
Marigold \texttt{v1.1} ($1{\times}1$)                 & 0.568 \\
Marigold \texttt{v1.1} ($1{\times}1$) (fp16)          & 0.274 \\
Marigold \texttt{v1.1} ($1{\times}1$) w/ TAESD (fp16) & 0.082 \\ 
\bottomrule
\end{tabular}

%% file: fig/4_qualitative_depth.tex
\newcommand{\sampleNameSingle}[1]{
    \multirow{1}{*}[1.15cm]{\rotatebox{90}{#1}}
}

\newcommand{\viscolumn}{0.125\textwidth}

\resizebox{\linewidth}{!}{
    \begin{tabular}[ht]{r p{\viscolumn} p{\viscolumn} p{\viscolumn} p{\viscolumn} p{\viscolumn} p{\viscolumn} p{\viscolumn} p{\viscolumn}}
        & \parbox[c]{\viscolumn}{\centering \ \ \ Input Image} 
        & \parbox[c]{\viscolumn}{\centering \ \ \raisebox{0.35ex}{MiDaS}} 
        & \parbox[c]{\viscolumn}{\centering \ \raisebox{0.35ex}{Omnidata}} 
        & \parbox[c]{\viscolumn}{\centering \raisebox{0.35ex}{DPT}} 
        & \parbox[c]{\viscolumn}{\centering \raisebox{0.35ex}{\nocompete{DAv2}}}
        & \parbox[c]{\viscolumn}{\centering \raisebox{0.35ex}{\nocompete{Metric3Dv2}}}
        & \parbox[c]{\viscolumn}{\centering{\!\!\!Marigold (ours)}} 
        & \parbox[c]{\viscolumn}{\centering \!\raisebox{0.35ex}{Ground Truth}} \\ 

        \sampleNameSingle{NYUv2}
        & \multicolumn{8}{c}{\includegraphics[width=\textwidth]{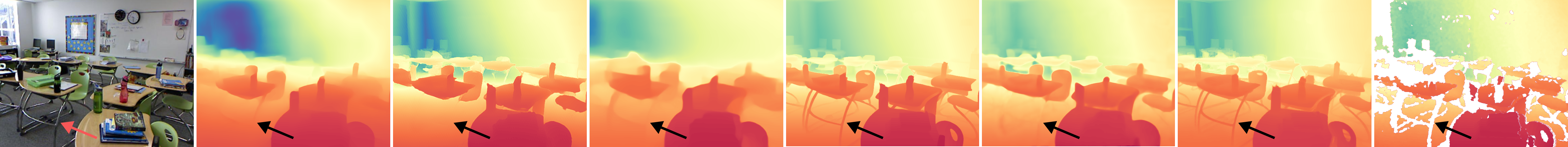}} 
        \\
        \sampleNameSingle{KITTI}
        & \multicolumn{8}{c}{\includegraphics[width=\textwidth]{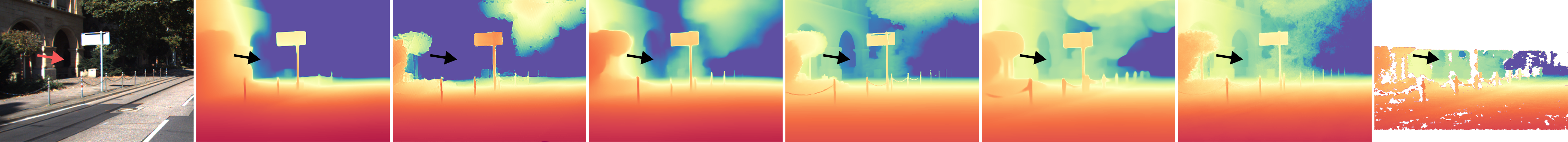}} 
        \\
        \sampleNameSingle{ETH3D}
        & \multicolumn{8}{c}{\includegraphics[width=\textwidth]{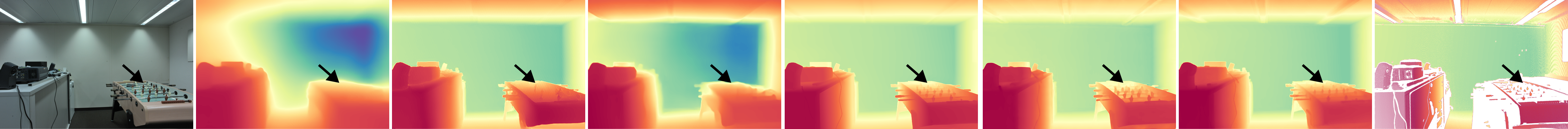}} 
        \\
        \sampleNameSingle{Scannet}
        & \multicolumn{8}{c}{\includegraphics[width=\textwidth]{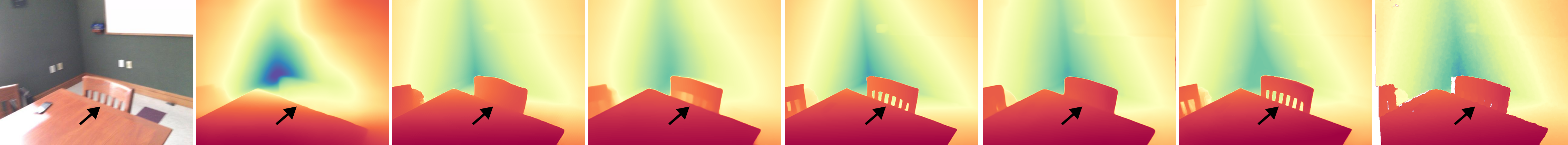}} 
        \\
        \sampleNameSingle{DIODE}
        & \multicolumn{8}{c}{\includegraphics[width=\textwidth]{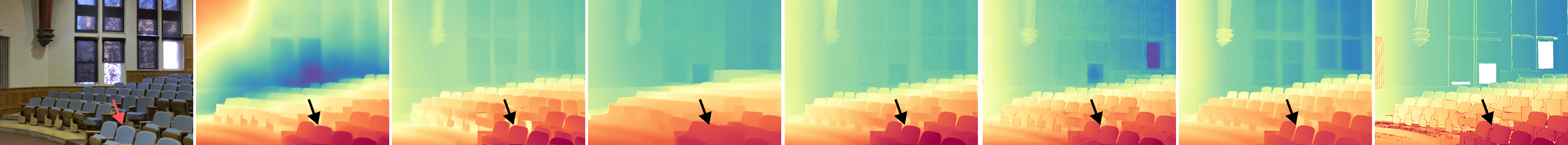}} 
    \end{tabular}
}

%% file: fig/5_qualitative_normal.tex
\newcommand{\sampleNameSingle}[1]{
    \multirow{1}{*}[1.3cm]{\rotatebox{90}{#1}}
}

\newcommand{\viscolumn}{0.166\textwidth}

\resizebox{\linewidth}{!}{
    \begin{tabular}[ht]{r p{\viscolumn} p{\viscolumn} p{\viscolumn} p{\viscolumn} p{\viscolumn} p{\viscolumn}}
        & \parbox[c]{\viscolumn}{\centering Input RGB Image} 
        & \parbox[c]{\viscolumn}{\centering MiDaS} 
        & \parbox[c]{\viscolumn}{\centering Omnidata} 
        & \parbox[c]{\viscolumn}{\centering DPT} 
        & \parbox[c]{\viscolumn}{\centering Marigold (ours)} 
        & \parbox[c]{\viscolumn}{\centering Ground Truth} \\ 
        \sampleNameSingle{NYUv2}
        & \multicolumn{6}{c}{\includegraphics[width=\textwidth]{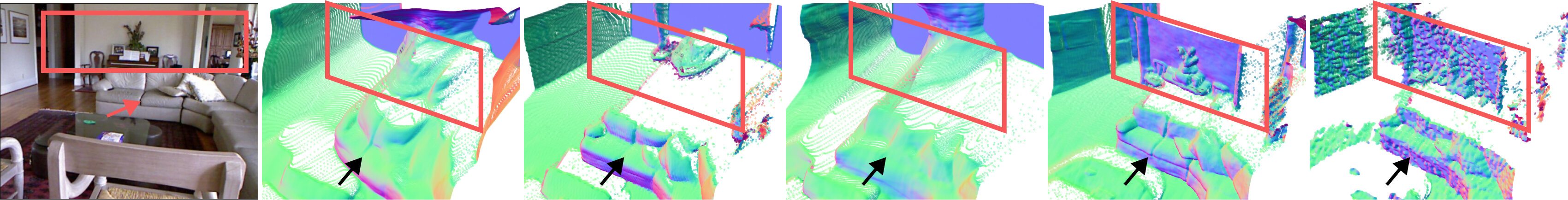}} 
        \\
        \sampleNameSingle{ScanNet}
        & \multicolumn{6}{c}{\includegraphics[width=\textwidth]{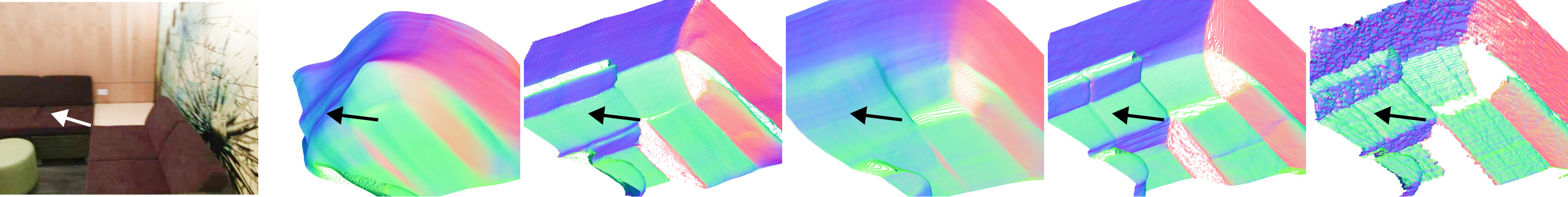}} 
        \\
        \sampleNameSingle{DIODE}
        & \multicolumn{6}{c}{\includegraphics[width=\textwidth]{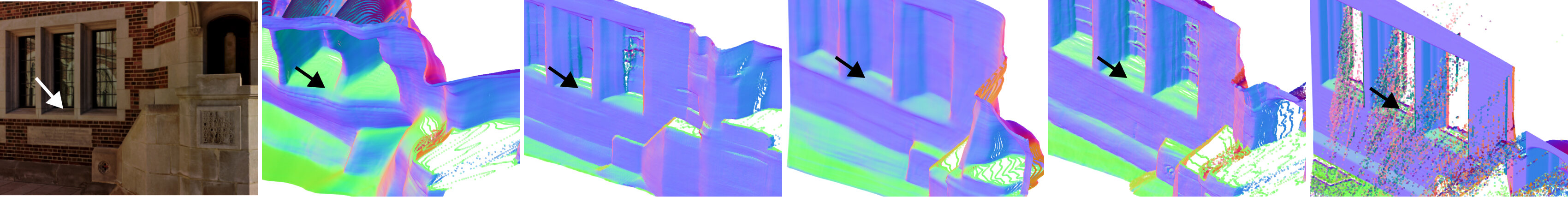}} 
    \end{tabular}
}

%% file: tbl/3_training_dataset.tex
\begin{tabular}{@{}cc c@{\hspace{3em}} cccc@{}}

\toprule

\multicolumn{1}{c}{\multirow{2}{*}{\makecell{HyperSim}}} & 
\multicolumn{1}{c}{\multirow{2}{*}{\makecell{Virtual \\KITTI}}} & 
& 
\multicolumn{2}{c}{NYUv2} & 
\multicolumn{2}{c}{KITTI} \\

\multicolumn{1}{c}{} & 
\multicolumn{1}{c}{} & 
& 
AbsRel ↓ & 
$\delta$1 ↑ & 
AbsRel ↓ & 
$\delta$1 ↑ \\

\midrule      

\xmark & \cmark &  & 13.9         & 83.4          & 15.4          & 79.3          \\
\cmark & \xmark &  & 5.7          & 96.3          & 13.7          & 82.5          \\
\cmark & \cmark &  & \textbf{5.6} & \textbf{96.5} & \textbf{11.3} & \textbf{88.7} \\

\bottomrule

\end{tabular}%

%% file: sec/3_method_normals.tex
\section{Surface Normals Estimation Model}
\label{sec:method:normals}

\begin{table*}[t!]
    \centering
    \caption{
        \textbf{Quantitative Comparison} of \method{}-Normals with SOTA surface normals estimators on several zero-shot benchmarks$^1$.
        In the 1$^{\mathrm{st}}$ section, we list 
        methods requiring more than 2M samples for fine-tuning.
        We compare methods from the 2$^{\mathrm{nd}}$ section with flavors of \method{}-Normals (ours, this paper) from the 3$^{\mathrm{rd}}$ section.
        Legend:
        The mean metric is presented as absolute angles, 11.25° metric is in percentage terms; 
        bold numbers are the best, underscored second best;
        NFEs is the number of function evaluations required to obtain the prediction -- $ensemble \times steps$ for diffusion models and $1$ for end-to-end networks.
        \method{} outperforms other methods in this low-data regime on most indoor and outdoor scenes.        
    }
    \resizebox{\linewidth}{!}{
	\input{tbl/5_Normals_quantitative}
	\label{table:normals_results}
    } 
    \\
    \begin{minipage}{0.98\linewidth}
        \scriptsize
        \vspace{0.4em}
        \begin{itemize}
        \item[$^1$]
            Metrics on ScanNet, NYUv2, and iBims-1 are sourced from the respective papers. 
            Metrics that were not reproducible are re-computed by us (for GeoWizard, GenPercept, and StableNormal). 
            We compute the metrics for all methods on OASIS (except for Omnidata and DSINE) and DIODE.
        \item[$^2$]
            Privileged information used by methods: 
            Metric3D v2 and DSINE require camera intrinsics; 
            GeoWizard requires choosing between indoor and outdoor regimes.
        \item[$^3$] 
            The preview models (\texttt{v0.1}), uploaded to the Hugging Face repository shortly after releasing \method{}-Depth (\texttt{v1.0})~\cite{ke2023repurposing}.
        \end{itemize}
    \end{minipage}
\end{table*}

Surface normals and depth estimation are inherently related, as both aim to regress 3D geometry. 
While the depth of a pixel is predicted as a single positive scalar, the surface normal is represented as a three-dimensional vector on the unit sphere.
Real surface normals can hardly be collected outside of a simulation or a controlled environment. 
Instead, normals have traditionally been derived from depth measurements, which often introduce noise at flat surfaces and unrealistic smoothness at depth discontinuities. 
Simulated data, however, often struggles with the sim-to-real gap.
This motivates \method{}-Normals, which aims to bridge the gap to real data through its Stable Diffusion prior.

\subsection{Method}\label{sec:normals_method}
We closely follow the \method{}-Depth fine-tuning protocol and introduce \method{}-Normals, a model variant for monocular surface normals estimation.
Methodology adaptation for the new task to cope with the three-dimensional unit vectors of surface normals is minimal. 
Raw ground-truth normal maps are streamed directly into the VAE encoder during training.
No range normalization or channel replication is required.
VAE outputs are normalized along the channel dimension to ensure unit-length predictions. 

\begin{figure}[t!]
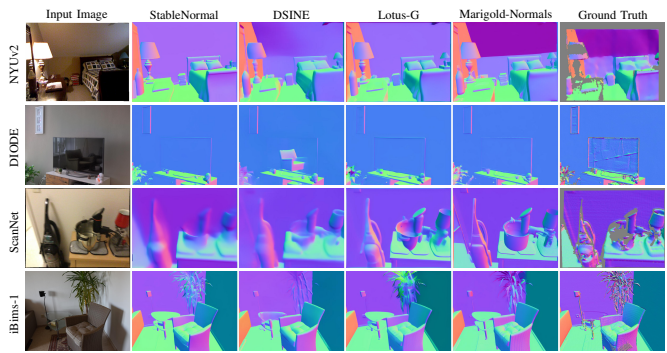

    \centering
    \setlength\tabcolsep{0.5pt}
        \include{fig/8_qualitative_surfaceNormals}
    \caption{
        \textbf{Qualitative comparison}
        of monocular surface normals estimation methods across different datasets. 
        Compared to baseline methods, \method{}-Normals demonstrates superior performance in handling complex scene layouts and shows greater robustness to motion blur and reflections.
    }
    \label{fig:normals_benchmark_visually}
\end{figure}

\vspace{0.35em}
\noindent\textbf{%
Test-time ensembling.
}
First, we run inference $N$ times with different noise initialization. 
We then average the normals predictions $\{ \mathbf{\hat{n}}_1, \ldots, \mathbf{\hat{n}}_N\}$ to a single mean prediction $\mathbf{\bar{n}}$ and normalize it to unit length. 
Lastly, for every pixel $(u,v)$ in the final prediction, we select the nearest neighbor vector from the $i$-th prediction $\mathbf{\hat{n}}^{(u,v)}_{i}$ that maximizes the cosine similarity with the corresponding vector in the mean normal map $\mathbf{\bar{n}}^{(u,v)}$:
    $\operatorname*{arg\,max}_{i \in \{1, \dots, N\}} \, \mathbf{\bar{n}}^{(u,v)} \cdot \mathbf{\hat{n}}^{(u,v)}_i$.

\subsection{Implementation} \label{sec:normals_implementation} 
We fine-tune the model for 26K iterations using the Adam optimizer with a learning rate of $6\cdot 10^{-5}$.
The training data is augmented with horizontal flipping, blurring, and color jitter. 
At inference, we use the DDIM scheduler~\cite{song2020denoising_ddim} in trailing mode and perform 4 steps. 
The final prediction is aggregated using an ensemble size of 10.
All other settings are the same as for the depth model.

\vspace{0.35em}
\noindent\textbf{Training Datasets.} 
We train the model on three synthetic datasets covering both indoor and outdoor scenes. 
Hypersim~\cite{roberts2021hypersim} and InteriorVerse~\cite{zhu2022interiorverse} are photo-realistic indoor datasets. We filter out incomplete samples and obtain 49K samples from 434 scenes for HyperSim and 27K samples from 3806 scenes for InteriorVerse. 
The training resolution is kept at $480 \times 640$ for both datasets. 
Sintel~\cite{Butler2012sintel} consists of indoor and outdoor sequences from a short animated film. 
We filter out low-quality samples and use 627 training images. 
The images are center-cropped to an aspect ratio of $3:4$ and resized to $480 \times 640$. 
Despite the small sample size, we observe improvement in training with Sintel due to its scene diversity and image appearance.

\begin{figure}[t]
    \centering
    \includegraphics[width=0.99\linewidth]{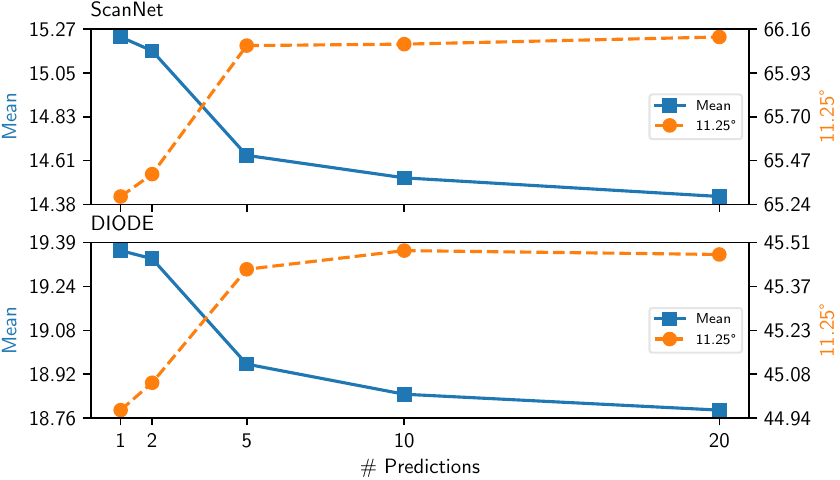}
    \caption{
        \textbf{Ablation of ensemble size} for \method{}-Normals.
        The performance consistently improves with increasing ensemble size. Diminishing returns begin after 10 predictions per sample.
    }
   \label{fig:normals_ablation_ensemble}
\end{figure}

\begin{figure}[t]
    \centering
    \includegraphics[width=0.99\linewidth]{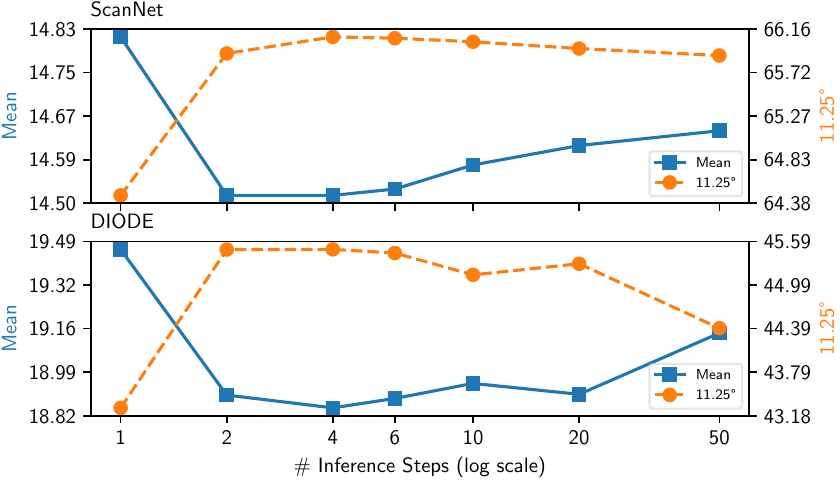}
    \caption{
        \textbf{Ablation of denoising steps} for \method{}-Normals.
        The best performance in benchmarks is obtained at 4 denoising steps. 
        Visually, one step is sufficient in most cases.
    }
   \label{fig:normals_ablation_steps}
\end{figure}

\subsection{Evaluation}\label{sec:normals_evaluation}
\vspace{0.35em}
\noindent\textbf{Evaluation Protocol.}
We evaluate our method on five unseen benchmarks. 
NYUv2~\cite{SilbermanECCV12nyu}, ScanNet~\cite{dai2017scannet}, and iBims-1~\cite{Koch2018ibims} are indoor depth datasets, for which we use the ground-truth normals provided by~\cite{bae2024dsine}. 
This includes 654 samples from the NYUv2 test split, all 100 samples of iBims-1, and a subset of 300 samples defined by~\cite{huang2019framenet} of ScanNet.
DIODE~\cite{diode_dataset} features both indoor and outdoor scenes. 
We use the validation split, encompassing 325 indoor and 446 outdoor samples.
OASIS~\cite{chen2020oasis} contains in-the-wild images sourced from the internet. 
We use the entire validation split of 10,000 samples.
Following established methods~\cite{Bansal16,Fouhey13,huang2019framenet}, we report the mean angular error and the percentage of pixels with an angular error \textless $11.25$°.

\vspace{0.35em}
\noindent\textbf{%
Comparison with other methods.
}
We compare \method{}-Normals to 8 baselines.  
DSINE~\cite{bae2024dsine} is a discriminative regression-based method that relies on a CNN architecture. 
Similar to us, GeoWizard~\cite{fu2024geowizard}, StableNormal~\cite{ye2024stablenormal}, and Lotus-G~\cite{he2024lotus} are generative diffusion-based methods that fine-tune a Stable Diffusion backbone. 
GenPercept~\cite{xu2024genpercept}, Lotus-D~\cite{he2024lotus}, and E2E-FT~\cite{garcia2024fine} bypass the probabilistic formulation and perform end-to-end training and inference instead.

The quantitative results are shown in Table~\ref{table:normals_results}. Our method consistently outperforms the baselines on most datasets and metrics. Notably, our straightforward fine-tuning and inference approach proves more effective than the more complex strategies employed by other diffusion-based methods. A qualitative comparison is shown in Figure~\ref{fig:normals_benchmark_visually}. 
It is apparent that \method{}-Normals produces highly accurate predictions, even in challenging scene layouts and scenarios. 

\subsection{Ablations}
\vspace{0.35em}
\noindent\textbf{Test-time ensembling.}
We investigate the impact of test-time ensembling while varying the number of aggregated predictions. 
The ablation studies are conducted on the ScanNet and DIODE splits.
As illustrated in Fig.~\ref{fig:normals_ablation_ensemble}, the performance with ensembling behaves very similarly to the depth model. 
While a single prediction already delivers solid results, the performance consistently improves as more predictions are combined. 
However, like depth, the gains plateau when more than 10 predictions per image are aggregated.

\begin{figure}[t!]
    \centering
    \includegraphics[width=0.99\linewidth]{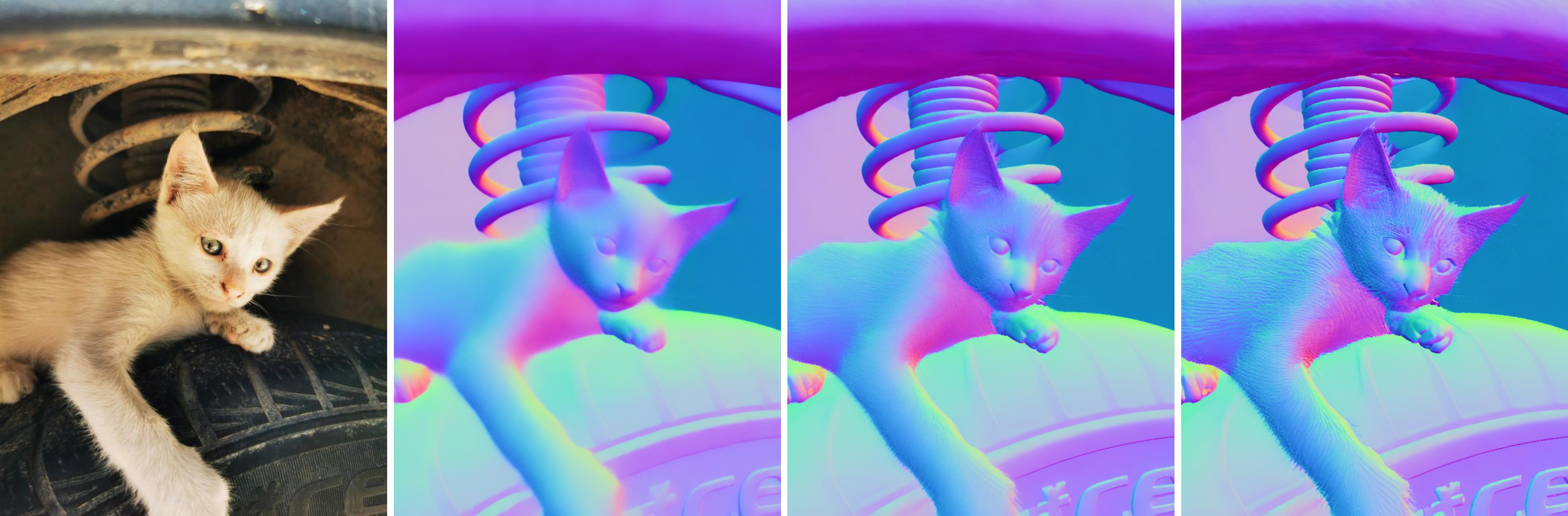}
    \caption{
        \textbf{Prediction granularity and denoising steps}. 
        From left to right, we visualize predictions with 1, 4, and 20 denoising steps during inference. 
        By increasing the number of steps, fine details, such as the cat's fur, become more pronounced.
    }
   \label{fig:normals_fine_details}
\end{figure}

\vspace{0.35em}
\noindent\textbf{Number of denoising steps.}
As quantitatively shown in Fig.~\ref{fig:normals_ablation_steps}, the best performance with the DDIM trailing timesteps setting is achieved at 4 steps. 
We additionally visualized predictions with 1-, 4-, and 20-step inference in Fig.~\ref{fig:normals_fine_details}. 
Interestingly, the level of detail can be controlled by simply adjusting the number of denoising iterations. 
While 1-step inference already produces reasonably good results in all cases, increasing the number of steps results in more pronounced details in high-frequency regions. 
However, improved details do not necessarily translate to improved performance metrics, as most evaluation benchmarks either mask out or over-smooth high-frequency regions in the ground truth. 

%% file: tbl/5_Normals_quantitative.tex
\begin{tabular}{
@{}l@{\hspace{1.5em}} @{} %
r@{\hspace{0.5em}} %
c@{\hspace{2em}} %
c@{\hspace{0.8em}} %
c@{\hspace{0.3em}} %
c @{}p{2.0em}@{} %
c@{\hspace{0.5em}}c @{}p{2.0em}@{} 
c@{\hspace{0.5em}}c @{}p{2.0em}@{} 
c@{\hspace{0.5em}}c @{}p{2.0em}@{} 
c@{\hspace{0.5em}}c @{}p{2.0em}@{} 
c@{\hspace{0.5em}}c 
}

\toprule

\multirow{2}{*}{Method} &
\multirow{2}{*}{NFEs} &
\multirow{2}{*}{Prior} &
\multicolumn{3}{c}{Data} & &
\multicolumn{2}{c}{ScanNet} & &
\multicolumn{2}{c}{NYUv2} & &
\multicolumn{2}{c}{iBims-1} & &
\multicolumn{2}{c}{DIODE} & &
\multicolumn{2}{c}{OASIS} 
\\

& 
& &
Prior & 
Real & 
Synthetic & &
Mean ↓ & 
11.25° ↑ & &
Mean ↓ & 
11.25° ↑ & &
Mean ↓ & 
11.25° ↑ & &
Mean ↓ & 
11.25° ↑ & &
Mean ↓ & 
11.25° ↑ 
\\

\midrule

\nocompete{Metric3D v2}~\cite{hu2024metric3d}\nocompete{$^2$} &
\nocompete{1} & 
\nocompete{DINOv2} &
\nocompete{142M} &
\nocompete{16M} & 
\nocompete{91K} & &  %
\nocompete{---} & 
\nocompete{---} & &
\nocompete{13.3} & 
\nocompete{66.4} & &
\nocompete{19.6} & 
\nocompete{69.7} & &
\nocompete{12.6} & 
\nocompete{64.9} & &
\nocompete{23.4} & 
\nocompete{28.5} 
\\

 \nocompete{Omnidata v2}~\cite{kar2022omnidata} & 
\nocompete{1} & 
\nocompete{\scriptsize ImageNet} &
\nocompete{15M} &
\nocompete{12M} & 
\nocompete{310K} & &
\nocompete{16.2} & 
\nocompete{60.2} & & 
\nocompete{17.2} & 
\nocompete{55.5} & &
\nocompete{18.2} & 
\nocompete{63.9} & &
\nocompete{20.6} & 
\nocompete{40.8} & &
\nocompete{24.2} & 
\nocompete{27.7} 
\\

\midrule

DSINE~\cite{bae2024dsine}$^2$ & 
1 & 
\scriptsize ImageNet &
1.3M &
86K &
74K & &
16.2 & 
61.0 & & 
16.4 & 
59.6 & &
17.1 & 
67.4 & &
19.9 & 
41.8 & &
24.4 & 
28.8
\\

GeoWizard~\cite{fu2024geowizard}$^2$ & 
$10 \times 50$ &
SD v2.0 &
2.3B &
51K &  %
227K & &  %
17.6 & 
54.6 & &
19.0 & 
50.0 & &
19.3 & 
62.3 & &
24.7 & 
30.1 & &
25.3 & 
26.9 
\\

GenPercept~\cite{xu2024genpercept} & 
1 &  %
SD v2.1 &
2.3B &
--- & 
44K & & 
18.2 & 
57.4 & & 
18.3 & 
56.0 & &
18.3 & 
63.8 & &
22.3 & 
38.1 & &
25.9 & 
23.3 
\\

StableNormal~\cite{ye2024stablenormal} & 
1 & 
SD v2.0 &
2.4B &
51K &  %
227K & &  %
16.7 & 
54.0 & &
17.8 & 
54.2 & &
17.1 & 
67.8 & &
19.3 & 
\textbf{53.8} & &
25.7 & 
25.4 
\\

Lotus-G~\cite{he2024lotus} & 
$1 \times 1$  &  %
SD v2.0 &
2.3B &
--- & 
59K & & 
15.3 & 
64.0 & & 
16.9 & 
59.1 & &
17.5 & 
66.1 & &
21.2 & 
39.7 & &
24.7 & 
27.0 
\\

Lotus-D~\cite{he2024lotus} & 
1 &  %
SD v2.0 &
2.3B &
--- & 
59K & & 
15.3 & 
62.9 & & 
16.8 & 
58.2 & &
17.7 & 
64.9 & &
21.0 & 
39.7 & &
25.7 & 
25.3 
\\

E2E-FT~\cite{garcia2024fine} & 
1 &  %
Marigold &
2.3B &
--- & 
74K & & 
\underline{14.7} & 
66.0 & & 
\underline{16.2} & 
\textbf{61.4} & &
\textbf{15.8} & 
\textbf{69.9} & &
19.2 & 
43.8 & &
\underline{22.8} & 
\underline{29.8}
\\

E2E-FT~\cite{garcia2024fine} & 
1 &  %
SD v2.0 &
2.3B &
--- & 
74K & & 
\underline{14.7} & 
\textbf{66.1} & & 
16.5 & 
60.4 & &
\underline{16.1} & 
\underline{69.7} & &
\underline{19.0} & 
44.4 & &
23.6 & 
27.9
 
\\

\midrule

\nocompete{Marigold-Normals \texttt{v0.1}$^3$} & 
$10 \times 50 $&
SD v2.0 &
2.3B &
--- & 
39K & &
\nocompete{16.1} & 
\nocompete{62.3} & &
\nocompete{17.1} & 
\nocompete{58.5} & &
\nocompete{16.6} & 
\nocompete{68.0} & &
\nocompete{19.6} & 
\nocompete{44.7} & &
\nocompete{23.5} & 
\nocompete{28.0} 
\\

Marigold-Normals \texttt{v1.1} & 
$1 \times 1$ &
SD v2.0 &
2.3B &
--- & 
77K & &
14.9 & %
64.3 & & %
16.4 & 
58.9 & &
17.2 & 
65.6 & &
19.5 & 
43.2 & &
23.2 & 
28.3

\\

Marigold-Normals \texttt{v1.1} & 
$10 \times 1$ &
SD v2.0 &
2.3B &
--- & 
77K & &
14.8 & %
64.5 & & %
16.3 & 
59.0 & &
17.1 & 
65.7 & &
19.5 & 
43.3 & &
23.2 & 
28.3

\\
Marigold-Normals \texttt{v1.1} & 
$1 \times 4$ &
SD v2.0 &
2.3B &
--- & 
77K & &
15.2 & %
65.3 & & %
17.0 & 
59.6 & &
17.0 & 
68.0 & &
19.4 & 
45.0 & &
23.2 & 
29.2

\\

Marigold-Normals \texttt{v1.1} & 
$10 \times 4$ &
SD v2.0 &
2.3B &
--- & 
77K & &
\textbf{14.5} & %
\textbf{66.1} & & %
\textbf{16.1} & 
\underline{60.5} & &
16.3 & 
68.5 & &
\textbf{18.8} & 
\underline{45.5} & &
\textbf{22.4} & 
\textbf{30.1} 
\\

\bottomrule

\end{tabular}

%% file: fig/8_qualitative_surfaceNormals.tex
\newcommand{\sampleNameSingle}[1]{
    \multirow{1}{*}[1.3cm]{\rotatebox{90}{#1}}
}

\newcommand{\viscolumn}{0.166\textwidth}

\resizebox{\linewidth}{!}{
    \begin{tabular}[ht]{r p{\viscolumn} p{\viscolumn} p{\viscolumn} p{\viscolumn} p{\viscolumn} p{\viscolumn}}
        & \parbox[c]{\viscolumn}{\centering Input Image} 
        & \parbox[c]{\viscolumn}{\centering StableNormal} 
        & \parbox[c]{\viscolumn}{\centering DSINE} 
        & \parbox[c]{\viscolumn}{\centering Lotus-G} 
        & \parbox[c]{\viscolumn}{\centering Marigold-Normals} 
        & \parbox[c]{\viscolumn}{\centering Ground Truth} \\ 

        \sampleNameSingle{NYUv2}
        & \multicolumn{6}{c}{\includegraphics[width=\textwidth]{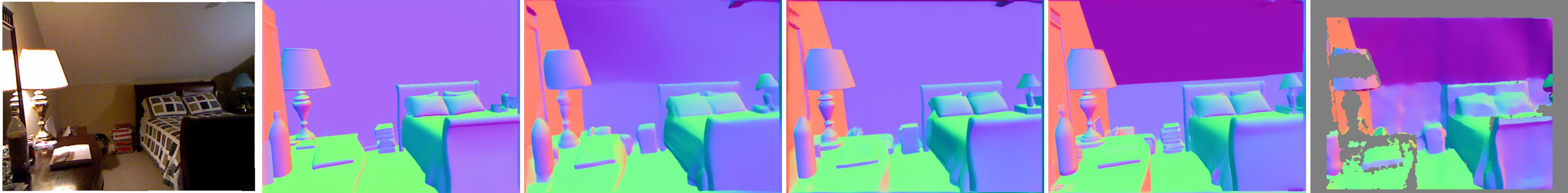}} 
        \\
        \sampleNameSingle{DIODE}
        & \multicolumn{6}{c}{\includegraphics[width=\textwidth]{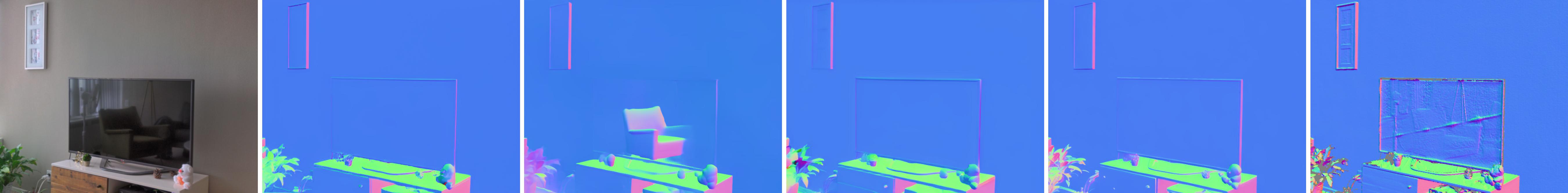}} 
        \\
        \sampleNameSingle{ScanNet}
        & \multicolumn{6}{c}{\includegraphics[width=\textwidth]{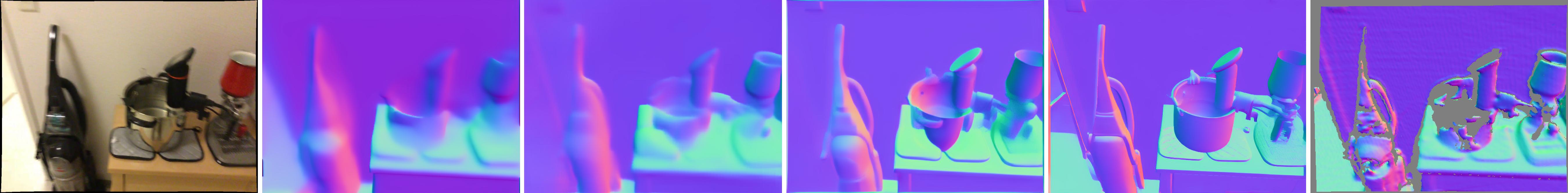}} 
        \\
        \sampleNameSingle{iBims-1}
        & \multicolumn{6}{c}{\includegraphics[width=\textwidth]{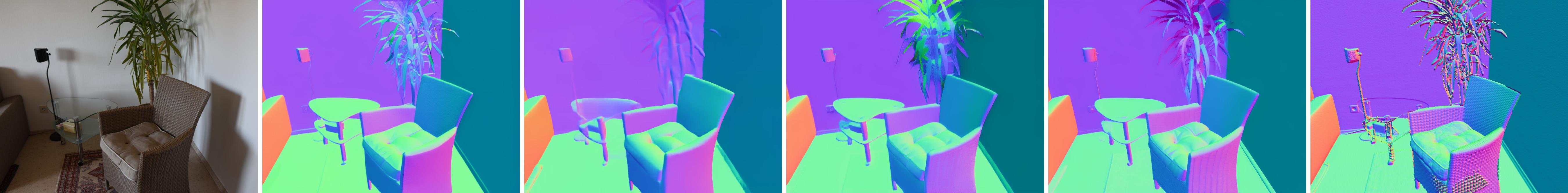}} 
    \end{tabular}
}

%% file: sec/3_method_albedo.tex
\section{Intrinsic Decomposition Models} 
\label{sec:method:iid}

Adaptation of \method{} to other image analysis tasks, such as Intrinsic Image Decomposition (IID), is simple and affordable. 
Specifically, this task enables a structured separation of images into physically meaningful properties. 
We introduce two models: \method{}-IID-Appearance and \method{}-IID-Lighting.

\method{}-IID-Appearance represents intrinsic properties through a physically-based BRDF, where material attributes are characterized by three key components: albedo, roughness, and metallicity. 
This model primarily focuses on estimating illumination-independent reflectance properties.

\method{}-IID-Lighting decomposes an image into albedo, diffuse shading, and a non-diffuse residual component. 
This decomposition aligns with the intrinsic residual model in linear space $I=A \cdot S + R$, where the image $I$ is composed of albedo $A$, a diffuse shading component $S$ (representing illumination color), and an additive residual term $R$ capturing non-diffuse effects. 
This formulation provides a structured way to separate reflectance from shading while accounting for complex illumination phenomena.

Similarly to the surface normals estimation task, ground truth for IID is hard to obtain without simulation or outside a controlled capture environment.
Therefore, we again turn to the available synthetic data to derive the \method{}-IID models.

\subsection{Method}
Different from depth and normals, IID requires predicting multiple images representing the decomposition of a single input image. 
In the case of \method{}-IID-Appearance, we predict two images per input: the first is the albedo image in standard color space, with values normalized to the unit range.
The second image encodes two BRDF properties (roughness and metallicity) into the red and green channels~\cite{kocsis2024intrinsic}, also normalized to a unit range. We denote this modality as \textit{material}. 
We keep the blue channel at zero, which gives the material visualization a red-green tone. 
For \method{}-IID-Lighting, three images are predicted per input: the albedo image, shading, and residual components, all normalized to the unit range.

\subsection{Implementation}
The U-Net is adapted to handle the increased number of predicted images ($P$): 2 and 3 for the IID-Appearance and IID-Lighting models, respectively.
The input channels of the first convolutional layer are replicated $P+1$ times; the whole weight tensor is divided by the replication factor to maintain the activations statistics.
The output channels of the final layer are replicated $P$ times without changing weights. 

The modified IID-Appearance U-Net is fine-tuned for 40K iterations on the training split of InteriorVerse, consisting of 45K samples at $480 \times 640$ resolution. 
Gamma correction and conversion from linear to sRGB space are applied to the input scene and target intrinsic images. 
The IID-Lighting U-Net is fine-tuned for 36K iterations on a pre-filtered training split of HyperSim, yielding 24K samples. 
All samples are resized to $480 \times 640$ and converted to sRGB space while keeping the target intrinsic images in linear space.

For quantitative evaluation, we perform 4 denoising steps without ensembling. 
Other settings remain the same as for the depth model.
As with other \method{} models, 1 step is sufficient qualitatively.

\begin{table}[t!]
    \centering
    \caption{
        \textbf{Quantitative Comparison} of \method{}-IID-Appearance \texttt{v1.1} with SOTA methods on the InteriorVerse test set. 
        \method{} outperforms the two competing methods on this benchmark.
    }
    \resizebox{\linewidth}{!}{
	\input{tbl/6_IntrinsicDecomp_quantitive}
	\label{table:intrinsic_decomp_quantitive}
    } 
    \\
\end{table}

\begin{table}[t!]
    \centering
    \caption{
        \textbf{Quantitative Comparison} of \method{}-IID-Lighting \texttt{v1.1} with SOTA methods on the HyperSim test set. 
        \method{} achieves competitive performance on this benchmark.
    }
    \resizebox{\linewidth}{!}{
	\input{tbl/7_IID_lighting_quantitative}
	\label{table:intrinsic_decomp_lighting_quantitive}
    } 
    \\
\end{table}

\begin{figure*}[t]
    \centering
    \resizebox{\linewidth}{!}{
        \includegraphics[width=0.48\linewidth]{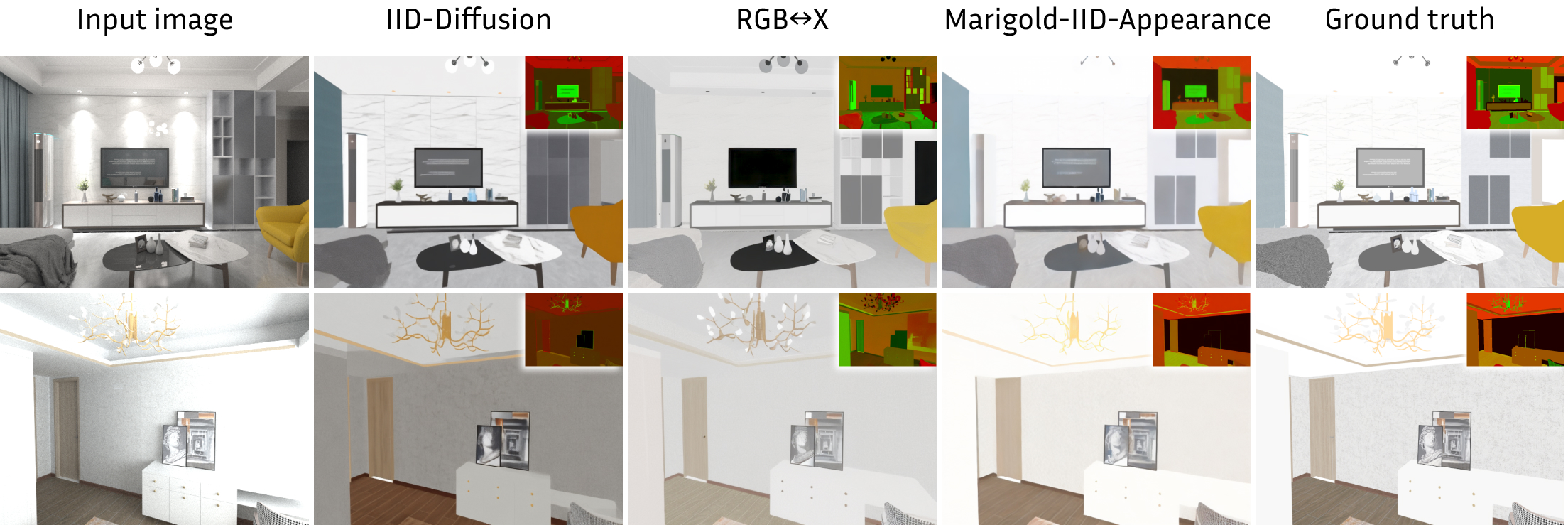}
        \ \ \ 
        \includegraphics[width=0.48\linewidth]{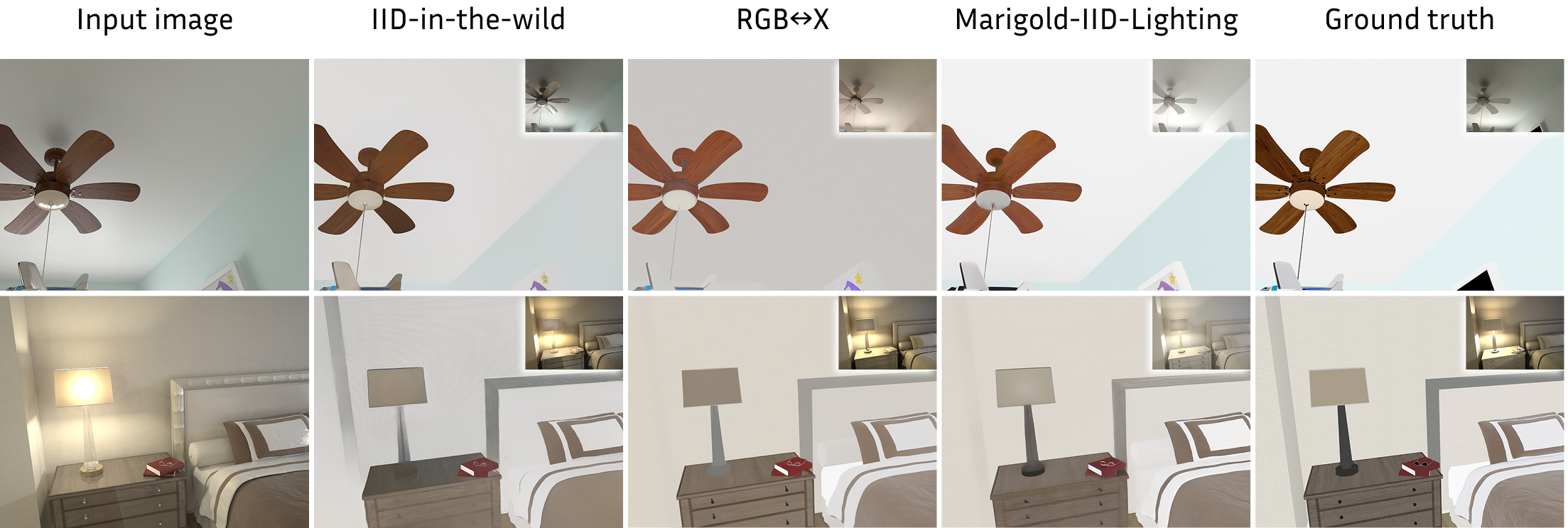}    
    }
    \caption{
        \textbf{Qualitative comparison} of \method{}-IID-Appearance (left) and \method{}-IID-Lighting (right).
        \textbf{Left}: albedo and material (with roughness in the red channel and metallicity in the green channel) predictions on the InteriorVerse test set.
        \textbf{Right}: albedo and diffuse shading predictions on the HyperSim test set.
        Predictions of \method{}-IID-Appearance and \method{}-IID-Lighting contain less baked-in shading and are more consistent with the ground truth.
    }
    \label{fig:albedo_comparison}
\end{figure*}

\begin{figure}[t!]
    \centering
    \includegraphics[width=0.99\linewidth]{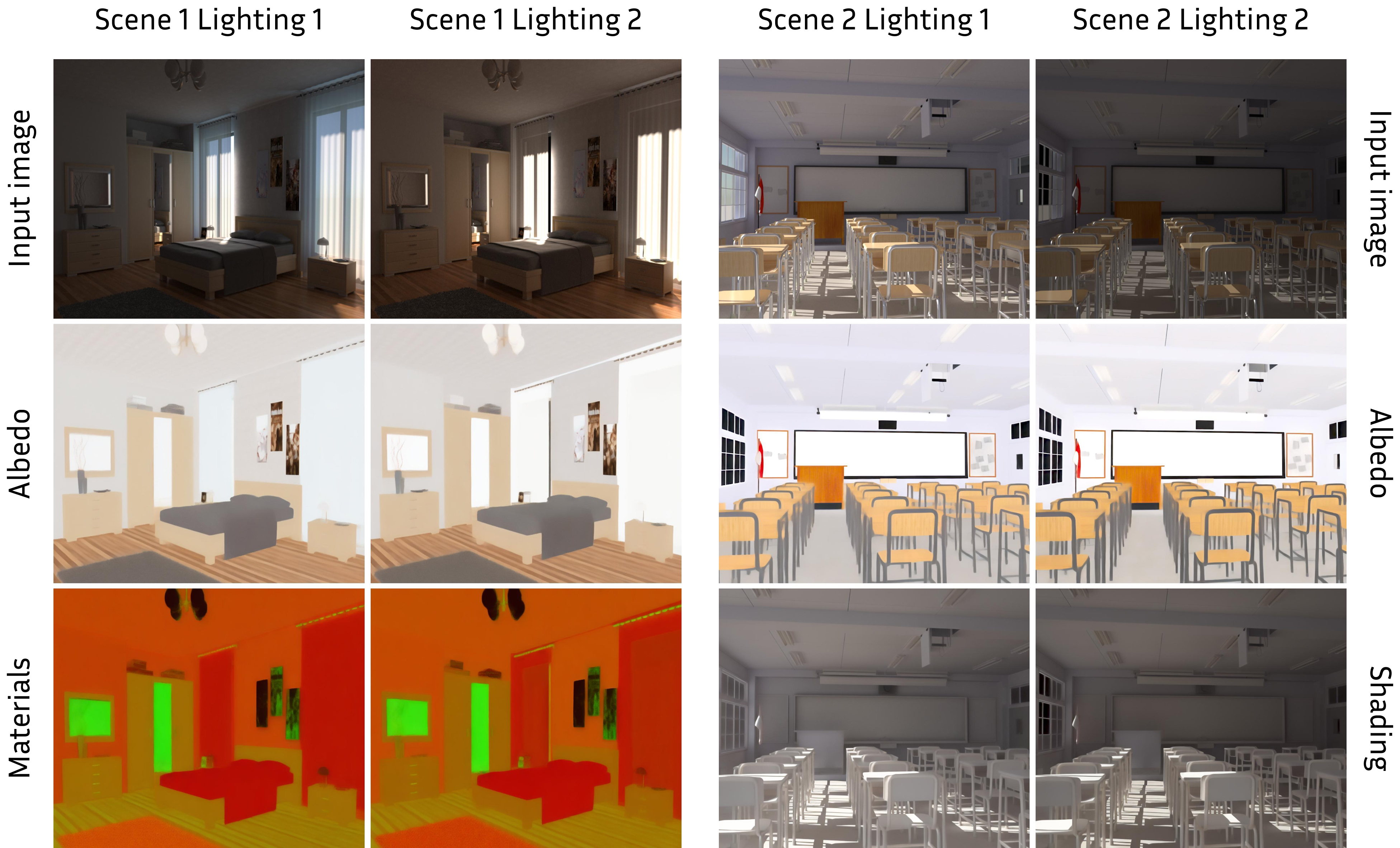}
    \caption{
        \textbf{Robustness to varying lighting conditions}. The \method{}-IID models generate consistent predictions across different environmental lighting setups of the same scene.
    }
   \label{fig:iid_varying_lighting}
\end{figure}

\begin{figure}[t!]
    \centering
    \includegraphics[width=0.99\linewidth]{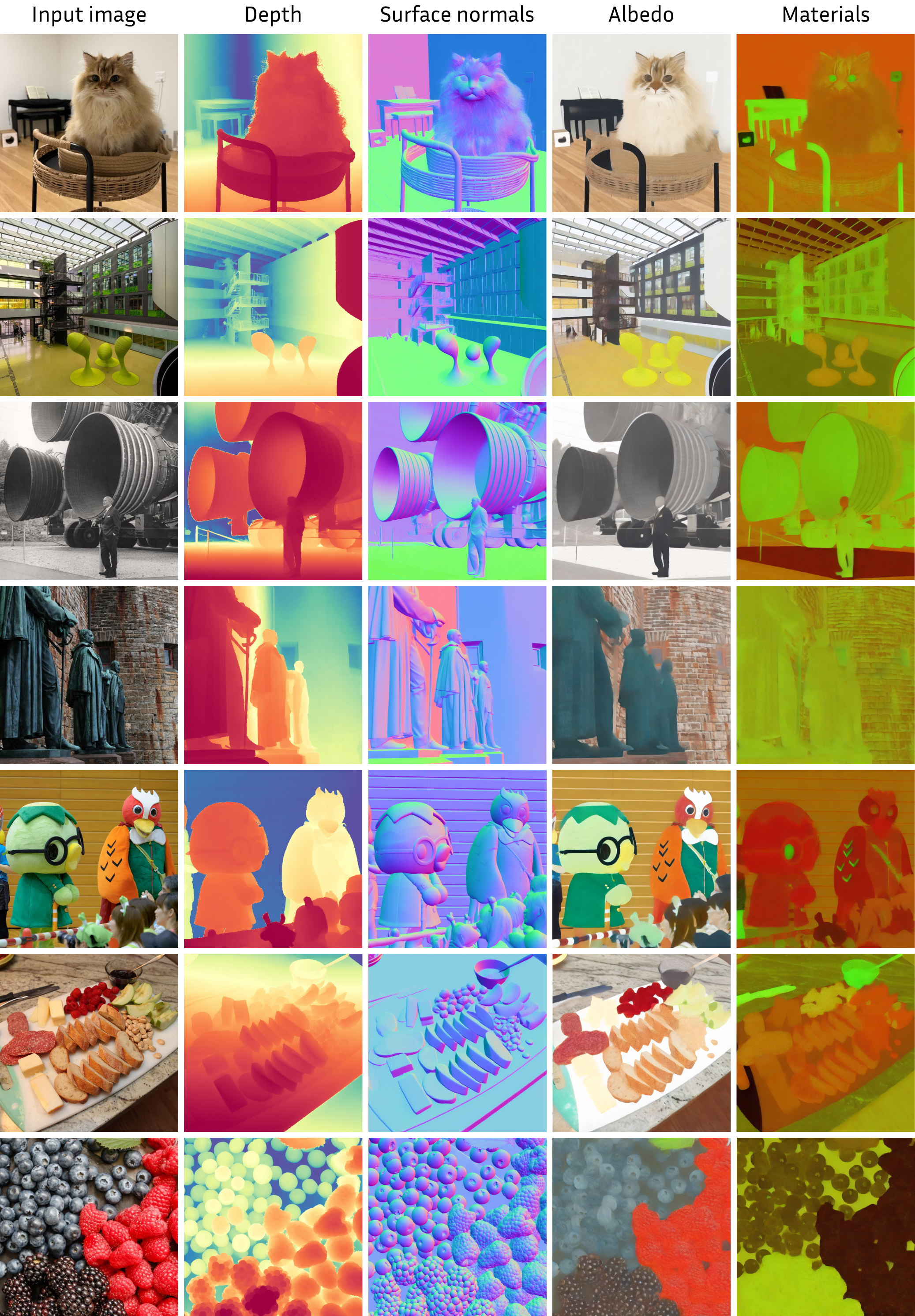}
    \caption{
        \textbf{\method{} in-the-wild results -- all modalities}.
        Our fine-tuning protocol enables generalization across multiple modalities. 
        None of the fine-tuning datasets included humans, animals, food, engines, or toys, attesting to the successful carry-over of the rich generative prior to downstream tasks.
    }
   \label{fig:gallery_all}
\end{figure}

\subsection{Evaluation}
\noindent\textbf{Evaluation Protocol.}
The IID-Appearance model is evaluated on the test split of the InteriorVerse~\cite{zhu2022interiorverse} dataset, which contains 2.6K samples. 
We assess the prediction performance of albedo and material. 
For IID-Lighting, evaluation is performed on the HyperSim~\cite{roberts2021hypersim} test split, comprising 5.2K samples, where we evaluate the quality of the predicted albedo and shading components.
Albedo and material predictions are compared directly to the ground truth without any alignment. 
Due to their differing value ranges, shading predictions are first scale-aligned to the corresponding ground truth and then normalized to the unit range before evaluation.
The reported metrics are Peak Signal-to-Noise Ratio (PSNR), Structural Similarity Index Measure (SSIM)~\cite{wang2004image}, and Learned Perceptual Image Patch Similarity (LPIPS)~\cite{zhang2018unreasonable}. 

\vspace{0.35em}
\noindent\textbf{Comparison with other methods}
We compare our results to three state-of-the-art methods: RGB$\leftrightarrow$X~\cite{zeng2024rgb} and Intrinsic Image Diffusion (IID-Diffusion)~\cite{kocsis2024intrinsic} for IID-Appearance, and RGB$\leftrightarrow$X and IID-in-the-wild~\cite{careaga2024colorful} for IID-Lighting.
For the cited methods, we adopt the inference settings reported in their respective papers: 50 denoising steps for RGB$\leftrightarrow$X, and 50 denoising steps with an ensemble size of 10 for IID-Diffusion.

The quantitative comparisons are shown in Tab.~\ref{table:intrinsic_decomp_quantitive} and Tab.~\ref{table:intrinsic_decomp_lighting_quantitive}. The visual comparisons are presented in Fig.~\ref{fig:albedo_comparison}.
Our method produces quantitatively more accurate and qualitatively cleaner decompositions of images. 
We further demonstrate the robustness of our method to varying environmental lighting conditions in Fig.~\ref{fig:iid_varying_lighting}.

\vspace{0.35em}
\noindent\textbf{All modalities in-the-wild} can be seen in Fig.~\ref{fig:gallery_all}. 
Evidently, image analysis tasks benefit from the rich generative prior, validating our universal fine-tuning protocol.

%% file: tbl/6_IntrinsicDecomp_quantitive.tex
\begin{tabular}{
@{}
l@{\hspace{0.9em}}
c@{\hspace{0.6em}}
c@{\hspace{0.6em}}
c@{\hspace{0.6em}}
c@{\hspace{0.7em}}
c@{\hspace{0.6em}}
c@{\hspace{0.6em}}
c@{\hspace{0.6em}}
@{}
}

\toprule

\multirow{2}{*}{Method} & 
\multicolumn{3}{c}{Albedo} &
& 
\multicolumn{3}{c}{Material}
\\ 

& 
PSNR $\uparrow$ & 
SSIM $\uparrow$ & 
LPIPS $\downarrow$ &
& 
PSNR $\uparrow$ & 
SSIM $\uparrow$ &
LPIPS $\downarrow$ \\

\midrule

IID-Diffusion~\cite{kocsis2024intrinsic} & 
\underline{18.10} & 
\textbf{0.863} & 
\underline{0.198} &
& 
\underline{16.09} & 
\underline{0.691} & 
\underline{0.356} 
\\ %

RGB$\leftrightarrow$X~\cite{zeng2024rgb} & 
13.16 & 
0.774 & 
0.289 &
& 
10.13 & 
0.547 & 
0.636 
\\ %

\midrule

Marigold-IID-Appearance \texttt{v1.1} & 
\textbf{19.50} & 
\underline{0.846} & 
\textbf{0.190} &
& 
\textbf{17.63} & 
\textbf{0.803} & 
\textbf{0.286} 
\\  %

\bottomrule

\end{tabular}

%% file: tbl/7_IID_lighting_quantitative.tex
\begin{tabular}{
@{}
l@{\hspace{0.9em}}
c@{\hspace{0.6em}}
c@{\hspace{0.6em}}
c@{\hspace{0.6em}}
c@{\hspace{0.7em}}
c@{\hspace{0.6em}}
c@{\hspace{0.6em}}
c@{\hspace{0.6em}}
@{}
}

\toprule

\multirow{2}{*}{Method} & 
\multicolumn{3}{c}{Albedo} &
& 
\multicolumn{3}{c}{Lighting}
\\ 

& 
PSNR $\uparrow$ & 
SSIM $\uparrow$ & 
LPIPS $\downarrow$ &
& 
PSNR $\uparrow$ & 
SSIM $\uparrow$ &
LPIPS $\downarrow$ \\

\midrule

IID-in-the-wild~\cite{careaga2024colorful} & 
\textbf{19.28} & 
\textbf{0.819} & 
0.260 &
& 
\underline{16.82} & 
0.725 & 
0.308 
\\ 

RGB$\leftrightarrow$X~\cite{zeng2024rgb} & 
17.43 & 
\underline{0.795} & 
\textbf{0.200} &
& 
16.70 & 
\textbf{0.742} & 
\textbf{0.251} 
\\ 

\midrule

Marigold-IID-Lighting \texttt{v1.1} & 
\underline{18.21} & 
0.771 & 
\underline{0.218} &
& 
\textbf{17.62} & 
\underline{0.729} & 
\underline{0.263}
\\  

\bottomrule

\end{tabular}

%% file: sec/3_method_lcm.tex
\section{Latent Consistency Model (LCM)}
\label{sec:method:lcm}

Latent Consistency Models~\cite{luo2023latent} is a latent diffusion model class that enables high-quality one- or few-step inference.
Inspired by LCM's success in fast image generation, we developed Marigold-LCM, a Latent Consistency Model variant of Marigold that achieves similar prediction results in one or a few denoising steps.

\label{sec:marigoldlcm}

\subsection{Method}

\begin{figure}[t!]
    \centering
    \includegraphics[width=\linewidth]{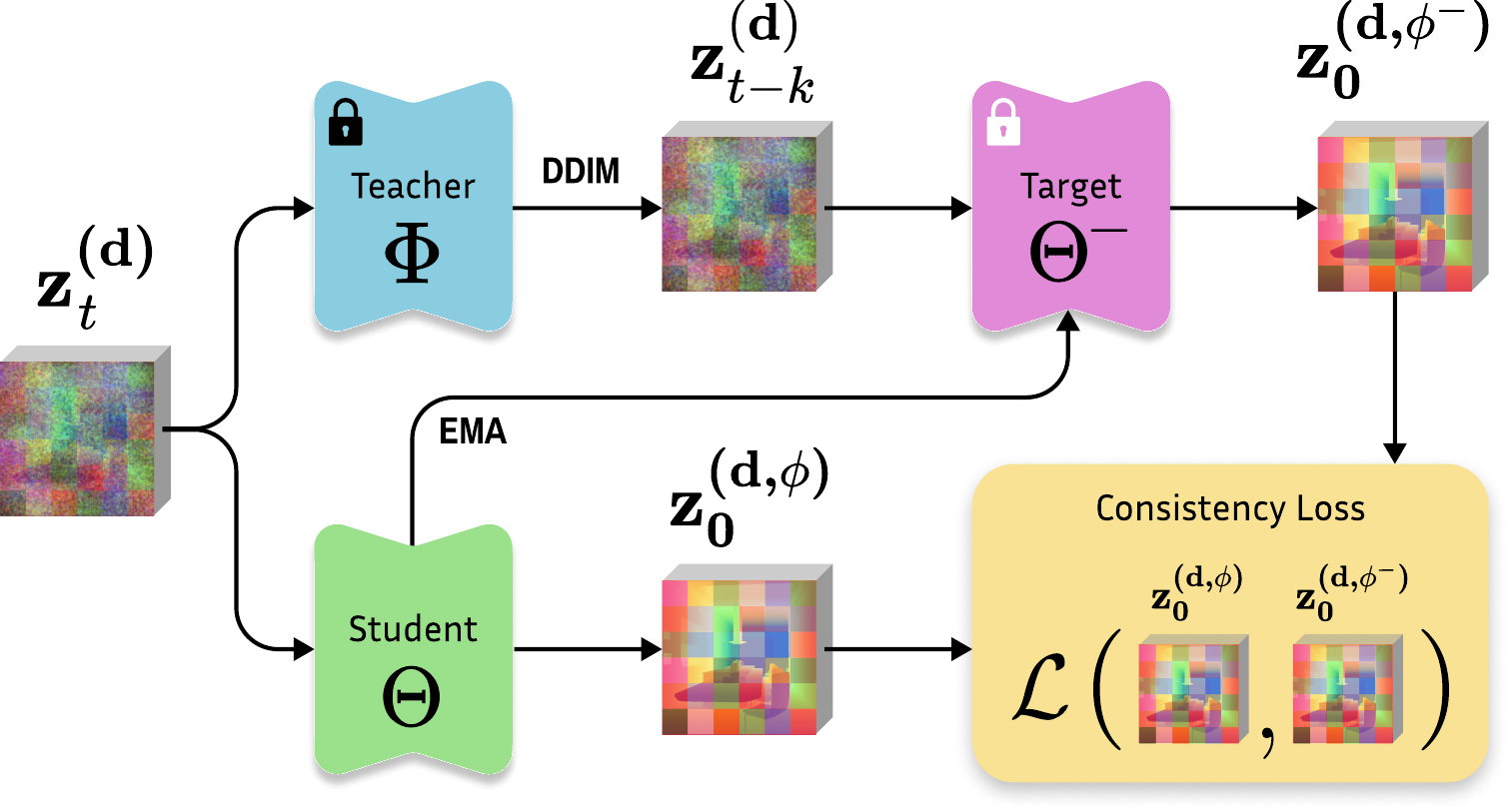}
    \caption{
    \textbf{Overview of Marigold-LCM distillation.}  
    To train Marigold-LCM, we initialize three replicas of the base \method{}: 
    the Teacher ($\Phi$), the Target ($\Theta^-$), and the Student ($\Theta$).
    The student is updated via optimization of the consistency objective, and the target is updated via EMA of student weights.
    At each training step, the student learns to predict the same clean latent $\mathbf{z^{(d)}_0}$ as produced by the teacher after applying the DDIM step of size $k$.
    Each model takes the image condition $\mathbf{z^{(x)}}$ and the input timestep, similarly to Fig.~\ref{fig:method-train}.
    }
    \label{fig:method-train-lcm}
\end{figure}

We distill Marigold-LCM from the standard Marigold using a similar recipe detailed in Luo~\etal~\cite{luo2023latent} (Fig.~\ref{fig:method-train-lcm}).
Similarly to the base fine-tuning protocol, we only distill the U-Net part of the model while keeping the VAE frozen.
The LCM distillation involves three models: 
a frozen \textit{teacher} model $\Phi$, which is a standard Marigold U-Net; 
a \textit{student} model $\Theta$, which we eventually output as Marigold-LCM;
and a \textit{target} model $\Theta^{-}$. 
Both the \textit{student} and \textit{target} models are initialized with the same weights as the teacher model.

At each training iteration, we sample data from the training dataset and convert the sample to latent images using the VAE encoder. 
We then sample Gaussian noise and diffuse the depth latent to a random timestep $t$ to obtain $\latentdepth_t$.
Now the \textit{teacher} model takes $\latentdepth_t$ as input, predicts noise, and then estimates the depth latent $\latentdepth_{t -k}$ at noise level $t-k$ using the DDIM solver step detailed in~\cite{luo2023latent}. 
We set $k=200$ in our implementation, which yields the best distillation result.
We now minimize the loss between the outputs of \textit{student} $\Theta$ and \textit{target} model  $\Theta^-$ through a self-consistency function $\mathbf{f}$~\cite{song2023consistency, luo2023latent}

\begin{equation}
    \mathcal{L} \left(\mathbf{f}(\Theta, \latentdepth_t, t), \mathbf{f}(\Theta^-, \latentdepth_{t-k}, t-k)\right).
\end{equation}

Here, the self-consistency function is defined as 
\begin{equation}
     \mathbf{f}(\theta, \mathbf{x}, t) = c_{\text{skip}}(t)  \mathbf{x} + c_{\text{out}}(t) \latent^{(\depth, \theta)}_0,
\end{equation}

where $c_{\text{skip}}, c_{\text{out}}$ are differentiable functions that satisfiy $c_{\text{skip}}(\epsilon) = 1, c_{\text{out}} = 0$ for some small $\epsilon > 0$,
and $\latent^{(\depth, \theta)}_0$ is the clean denoised depth latent predicted by model $\theta$.
We use the Pseudo-Huber metric~\cite{song2023improved} as our loss function:
\begin{equation}
     \mathcal{L}(\mathbf{x}, \mathbf{y}) = \sqrt{\|\mathbf{x} - \mathbf{y}\|_2^2 + c^2} - c,
\end{equation}
where we set $c = 0.001$.

At the end of each training iteration, the \textit{student} model is updated using gradient descent according to our loss function. 
While the weights of the \textit{target} model are updated with a running average of the weights of the \textit{student} model:
\begin{equation}
    \Theta^- = \text{stopgrad} \left(\mu \Theta^- + \left(1-\mu\right) \Theta\right),
\end{equation}
here the decay weight $\mu$ is set to be 0.95.

\subsection{Inference}

The consistency distillation training allows only one forward pass of the noise latent through the trained student model to generate the clean predicted depth latent. 
Additionally, one can improve the sample quality by taking multi-step sampling~\cite{luo2023latent, song2023consistency}.
This is done by adding another random noise latent to the denoised latent following a timestep schedule and then denoising again through the student model. 
In practice, we use the identical noise schedule of the original \method{}, swap the U-Net with the trained LCM model, and change the denoising sampling method from DDIM to LCM, as introduced above.
This means that the evaluation protocol of Marigold-LCM is identical to the base Marigold.

\subsection{Implementation}

The Marigold-LCM model is distilled from the base model for 5K iterations using the AdamW~\cite{loshchilov2017decoupled} optimizer with a base learning rate of $3\cdot 10^{-6}$. 
We use the same training set, batch size, gradient accumulation steps, and data augmentation as Marigold training. 
Distilling Marigold LCM takes approximately one day on a single NVIDIA A100 GPU card with 40G VRAM.
We apply one-step denoising at inference time to output the clean predicted depth latent. 
For evaluation, we follow the same ensemble method as the standard Marigold using 10 samples. 

\subsection{Experiments}

We compared Marigold-LCM with one LCM inference step with various Marigold DDIM configurations in Tab.~\ref{table:zeroshot_test}.
Although Marigold-LCM with one step does not outperform the original Marigold with $50$ steps in most cases, it outperforms prior art on most datasets and metrics. 
This validates the hypothesis that Marigold is amenable to the latent consistency distillation, and the resulting model is on par with the base.
It also shows that LCM distillation can be successfully adapted to modalities other than text-to-image.
However, given the improved quantitative and qualitative performance of Marigold with DDIM and trailing timesteps pointed out in E2E-FT~\cite{garcia2024fine}, the viability of LCM distillation remains an open question for future research.

%% file: sec/3_method_hr.tex
\section{High Resolution Depth Model} 
\label{sec:method:hr}

\begin{figure}[tp]
    \centering
    \includegraphics[width=1.0\linewidth]{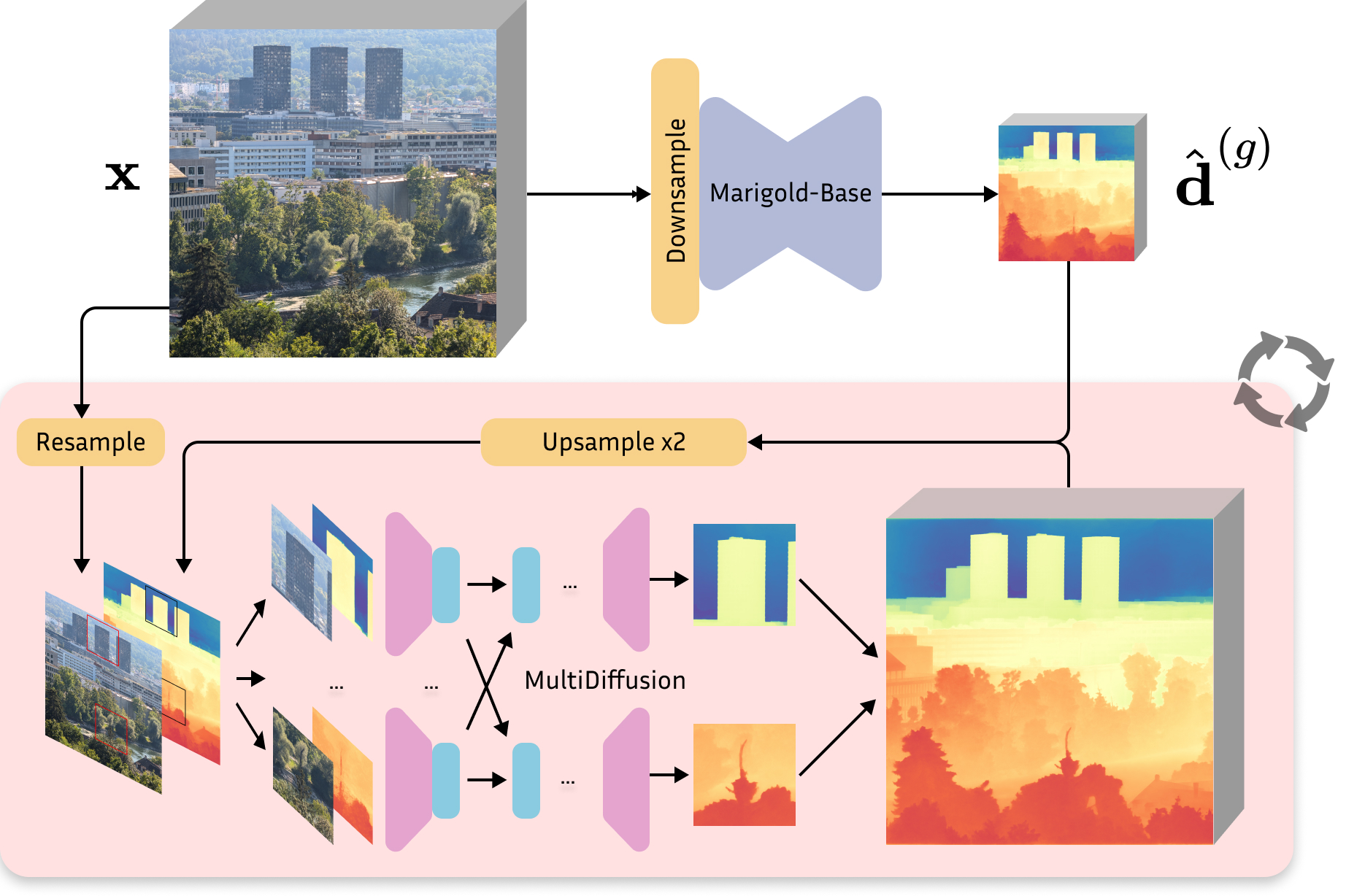}
    \caption{
    \textbf{High-resolution Marigold Pipeline.} 
    We first create a global prediction $\hat{\mathbf{d}}^{(g)}$ with the original \method{}-Depth pipeline at the native processing resolution. 
    This prediction is then used as an additional conditioning variable in the upsampling diffusion process, which upsamples the prediction in a patch-based, MultiDiffusion forward pass.
    }
    \label{fig:highres_meth}
\end{figure}

Applying monocular depth estimation networks to high-resolution images seems straightforward, but poses two inherent challenges. 
First, neural networks have a fixed receptive field limited to the model architecture or the resolution of the training data. 
Second, memory consumption can become excessive when naively applying those models to larger image dimensions.
The simple way to address those challenges is to downsample to its native processing resolution (768 for Marigolds fine-tuned from Stable Diffusion) and later upsample the computed depth map. 
We refer to this prediction as a ``global depth map'' $\hat{\mathbf{d}}^{(g)}$.
Typically, this approach compromises the edge quality of the depth estimation for high-resolution predictions. 
An alternative approach is partitioning a large image into smaller patches and processing them independently.
However, even if consistency at the seams was perfect, this method suffers from global inconsistencies due to the lack of communication between neighboring patches \wrt global layout.

\subsection{Method}

We introduce the \method{}-HR model to overcome these challenges (Fig.~\ref{fig:highres_meth}). 
The process starts by predicting a global depth map at native processing resolution, as in the original version. 
This global depth map serves as a coarse scene representation and is the initialization for the following refinement procedure.

Next, we upsample the global depth map by a factor $2\times$ and use it with the correspondingly resampled RGB image as the conditioning for another diffusion model named $\Phi$. 
To keep memory usage bounded, we implement the forward pass as a bundle of inferences of overlapping tiles, where we synchronize latents via a closed-form equation from the MultiDiffusion approach~\cite{bar2023multidiffusion}:
\begin{equation}
\Psi\left(J_t \mid z\right)=\sum_{i=1}^n \frac{F_i^{-1}\left(W_i\right)}{\sum_{j=1}^n F_j^{-1}\left(W_j\right)} \otimes F_i^{-1}\left(\Phi\left(\catinput_t^i\right)\right)
\end{equation}

$W_i$ are the per-pixel blending weights of tile $i$ -- in our implementation, we use the Chamfer distance to the image border. 
The function $F_i^{-1}$ transforms the tile back into its location in the global canvas based on the tile index $i$. For this model latent variable $\catinput_t^i$ is defined as the $F_i(\text{cat}(\latent^{(\depth)}_t, \latent^{(g)}_t, \latent^{(\img)}))$.

\subsection{Implementation}

For the \method{}-HR refiner, we resume the training from the \method{}-Depth checkpoint for 12K iterations with additional conditioning on the $\times 2$ lower resolution inference. 
We follow the BetterDepth protocol~\cite{zhang2024betterdepth}, first aligning the global prediction to the ground truth and then masking out the loss of dissimilar patches, such that the refiner is encouraged to follow the conditioning. 
We use the suggested setting of $\eta=0.1$ to threshold the distance between patches. 
For the MultiDiffusion pipeline, we set the patch overlap to 50\%.

We keep the same training datasets as the base model.
However, we deviate from the base protocol and train with half-resolution crops of size $384 \times 512$.
To generate the global conditioning, we precompute \method{}-Depth results for the whole training dataset with half the dataset resolution as the processing resolution. 
Specifically, we generate two sets of predictions: one set of base predictions with 10 ensemble members and one set using ensemble size one. 
We randomly select one of these sets for each sample during training. 
For data augmentations, we apply Gaussian blur to 50\% training samples with random radii ranging from 0 to 4 pixels.

\subsection{Evaluation}

\vspace{0.35em}
\noindent\textbf{%
Evaluation datasets.
}
We benchmark on two high-resolution datasets that contain unmasked depth continuities.
The datasets are: 
First, the stereo-matched Middlebury 2014~\cite{scharstein2014high} contains ground truth of resolution $2016 \times 2940$. 
We evaluate the complete dataset with 46 samples. 
Second, we evaluate the Booster dataset~\cite{ramirez2024ntire}, which contains a stereo-generated ground-truth of size $3008 \times 4112$. 
We only consider the scenes where the ground truth is provided. 
Since the images in each scene only vary by illumination and small parallax, we use one image per scene -- 31 samples total. 

\vspace{0.35em}
\noindent\textbf{%
Evaluation protocol.
}
We evaluate all depth maps with the affine-invariant protocol and provide the corresponding general-purpose metrics -- i.e., the Absolute Mean Relative Error (AbsRel~↓) and Threshold Accuracy ($\delta1$~↑) accuracy. 
Furthermore, four edge-based metrics evaluate the quality of the discontinuities: Depth Boundary Error Completeness ($\epscompdbe$~↓) and Accuracy ($\epsaccdbe$~↓)~\cite{koch2018evaluation}, as well as Edge Precision ($\edgeprc$~↑) and Recall ($\edgerec$~↑)~\cite{hu2019revisiting}. 
Since none of the methods directly outputs the exact dataset-specific resolutions, we resample the prediction using bilinear interpolation. 
For ours, we employ two upsampling iterations with Marigold-HR, until the output resolution approximates the target resolution.

\vspace{0.35em}
\noindent\textbf{Comparison with other methods.} 
\input{fig/6_HighResolution}
\begin{table*}[t!]
    \centering
    \caption{
        \textbf{Quantitative comparison} of \method{}-HR \texttt{v1.0} against SOTA depth estimators. We note the bootstraped model in the brackets after the actual method name. Marigold-HR improves upon the base model, particularly excelling in edge quality metrics, where it achieves results competitive with current state-of-the-art models. 
    }
    \resizebox{\linewidth}{!}{
	\input{tbl/4_HR_quantitative_v2}
	\label{table:HR_test}
    } 
\end{table*}
We benchmark our method against other recent methods designed for high-resolution inference. BoostingDepth~\cite{miangoleh2021boosting} implements a version based on MiDaS~\cite{Ranftl2020_midas} and another based on LeRes~\cite{Wei2021CVPR_leres}. 
We note that the officially provided checkpoints are trained on a mixed dataset, including our evaluation dataset, Middlebury 2014. 
Finally, we also compare the recent PatchFusion~\cite{li2024patchfusion} that bootstraps ZoeDepth~\cite{bhat2023zoedepth} and the concurrently proposed DepthPro~\cite{appledepthpro}.

In Table~\ref{table:HR_test}, we present the results on the high-resolution dataset.
\method{}-HR achieves the best or second-best performance in all metrics.
Depth Pro performs best in global depth estimation metrics (AbsRel and $\delta1$). In terms of edge quality metrics ($\epscompdbe$, $\epsaccdbe$, $\edgeprc$, $\edgerec$), Depth Pro is also a strong performer; however, on the Booster dataset, \method{}-HR achieves slightly better performance.

Furthermore, we show qualitative results in Fig.~\ref{fig:HR_viz}. 
These results demonstrate the visual quality attainable with a diffusion-based model at high resolution. 
For the in-the-wild example as well, our model produces plausible results, capturing even fine-grained details such as the cat’s whiskers.

\subsection{Ablations}
We conduct ablation studies to evaluate the impact of our model's two main methodological design choices: global conditioning and MultiDiffusion inference. The results are presented in Table~\ref{table:HR_ablations}.

Starting with the base \method{}-Depth model, we first augment it with global conditioning, which requires retraining the model. 
This modification improves the edge quality metrics, while the global metrics only slightly worsen. 
However, we note that GPU memory consumption scales quadratically with the upsampling factor in this configuration, making it less practical for high-resolution applications. 
Next, we apply the MultiDiffusion inference strategy to the \method{}-Depth model without retraining. 
This approach also improves the edge quality metrics at higher resolutions but keeps GPU memory bounded. 
However, we observe a degradation in the global metrics because the patches processed during inference lack global context. 
Finally, by combining global conditioning and MultiDiffusion inference, we improve global and edge-focused metrics as shown in the last two rows of Table~\ref{table:HR_ablations}; 
effectively balancing the benefits of global context and high-resolution edges, with moderate memory costs below 15GB.

\begin{table}[t!]
    \centering
    \caption{
        \textbf{Ablation study for Marigold's high-resolution module}. 
        Combining both global conditioning and MultiDiffusion inference yields the best overall performance compared to its ablated versions. 
    }
    \resizebox{\linewidth}{!}{
	\input{tbl/4_HR_quantitative_ablations}
	\label{table:HR_ablations}
    } 
    \\
    \begin{minipage}{0.98\linewidth}
        \scriptsize
        \vspace{0.4em}
        \begin{itemize}
        \item[$^{\ddag}$]
         denotes that inference had to be done in float16 to reduce GPU memory usage.
        \end{itemize}
    \end{minipage}
\end{table}

%% file: fig/6_HighResolution.tex
\begin{figure}[t]
    \centering
    \resizebox{\linewidth}{!}{
    \includegraphics[width=1.0\textwidth]{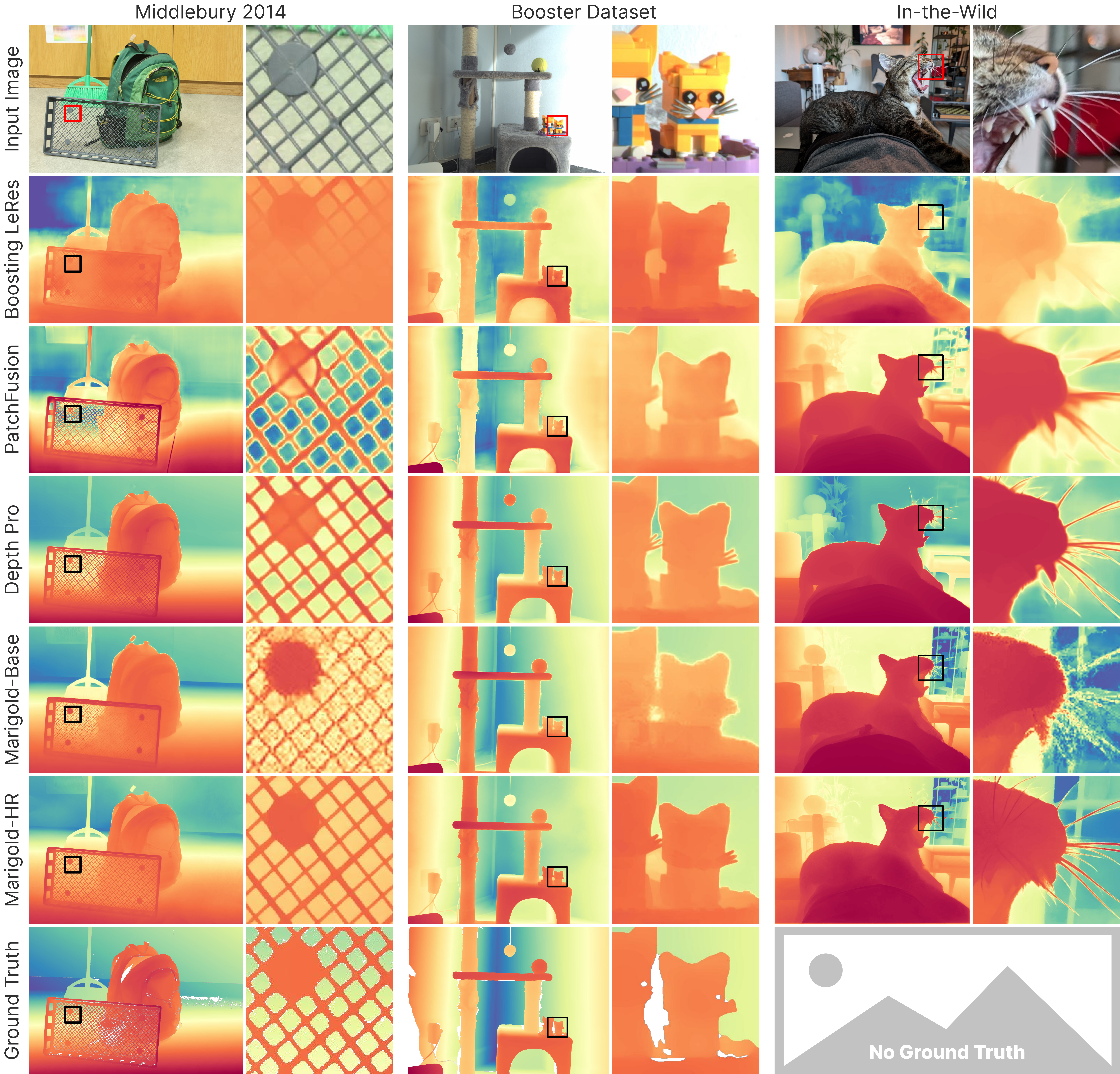}
    }
    \caption{
    \textbf{Quantitative comparison} of \method{}-HR, \method{}, and other SOTA methods.
    Predictions on Middlebury 2014 and Booster are aligned to the ground truth and visualized with the same color mapping. The predictions for the in-the-wild example are per sample normalized. \method{}-HR produces fine-grained outputs while also maintaining global context.}
    \label{fig:HR_viz}
\end{figure}

%% file: tbl/4_HR_quantitative_v2.tex
\begin{tabular}{
    @{}l
    c@{\hspace{0.5em}} %
    c@{\hspace{0.5em}} %
    c@{\hspace{0.5em}}
    c@{\hspace{1.0em}}
    c@{\hspace{0.5em}}
    c@{\hspace{1.0em}}
    c@{\hspace{0.5em}}
    c@{\hspace{0.5em}}
    c@{\hspace{0.5em}}
    c@{\hspace{0.5em}}
    c %
    c@{\hspace{0.5em}}
    c@{\hspace{1.0em}}
    c@{\hspace{0.5em}}
    c@{\hspace{0.5em}}
    c@{\hspace{0.5em}}
    c@{\hspace{0.5em}}
}

\toprule
\multirow{2}{*}{\vspace{-0.5em}Method} & 
\multicolumn{2}{c}{Data} & &
\multicolumn{6}{c}{Middlebury 2014}  & & 
\multicolumn{6}{c}{Booster Dataset} 
\\[0.5em]

& Real & Synthetic & &
AbsRel ↓ &
$\delta$1 ↑ &
$\epscompdbe$ ↓ &
$\epsaccdbe$ ↓ &
$\edgeprc$ ↑ &
$\edgerec$ ↑ & & 
AbsRel ↓ &
$\delta$1 ↑ &
$\epscompdbe$ ↓ &
$\epsaccdbe$ ↓ &
$\edgeprc$ ↑ &
$\edgerec$ ↑ 
\\

\midrule
Boosting~\cite{miangoleh2021boosting} (MiDaS~\cite{Ranftl2020_midas}) & 
2M & --- & &
7.3 & 94.2 & 5.9 & 1.9 & 30. & 21.  & & 
7.5 & 95.2 & 8.1 & 2.6 & 43. & 12.  
\\

Boosting~\cite{miangoleh2021boosting} (LeRes~\cite{Wei2021CVPR_leres}) & 
300K & 54K & &
7.0 & 95.3 & 5.4 & 1.9 & 29. & 29. & & 
6.9 & 94.7 & 6.2 & \underline{2.2} & \underline{44.} & 25. 
\\

PatchFusion~\cite{li2024patchfusion} (ZoeDepth~\cite{bhat2023zoedepth}) & 
3.9M & 1M  & &
6.4 & 96.2 & 4.5 & 1.9 & 29. & 42.  & & 
6.7 & 94.6 & 5.7 & 2.4 & \underline{44.} & 28.  
\\ 

Depth Pro~\cite{appledepthpro} & 
5.1M & 2.5M & &
\textbf{3.1} & \textbf{99.3} & \underline{3.6} & \textbf{1.7} & \textbf{36.} & \underline{54.} & & 
\textbf{2.0} & \textbf{99.8} & \underline{4.9} & 2.3 & \textbf{48.} & \underline{38.} 
\\

\midrule

Marigold-Depth \texttt{v1.0} & 
--- & 74K & &
\underline{5.0} & 97.5 & 5.0 & 2.4 & 24. & 29. & & 
\textbf{3.9} & 99.1 & 6.4 & 3.1 & 33. & 22. 
\\

Marigold-HR \texttt{v1.0} & 
--- & 74K & &
\underline{5.0} & \underline{97.8} & \textbf{3.5} & \underline{1.8} & \underline{33.} & \textbf{61.} & & 
\underline{4.3} & \underline{99.2} & \textbf{4.1} & \textbf{1.8} & \textbf{48.} & \textbf{45.} 
\\

\bottomrule

\end{tabular}

%% file: tbl/4_HR_quantitative_ablations.tex
\begin{tabular}{
    @{}l
    c
    c@{\hspace{0.5em}}
    c@{\hspace{1.0em}}
    c@{\hspace{0.5em}}
    c@{\hspace{0.5em}}
    c
    c@{\hspace{0.5em}}
    c@{\hspace{1.0em}}
    c@{\hspace{0.5em}}
    c@{\hspace{0.5em}}
    @{}
}

\toprule
\multirow{2}{*}{\vspace{-0.5em}Method} & 
\multirow{2}{*}{\vspace{-0.5em}Size} &
\multicolumn{4}{c}{Middlebury 2014} & &
\multicolumn{4}{c}{Booster Dataset} \\[0.5em]

& &
$\delta$1 ↑ &
$\epscompdbe$ ↓ &
$\epsaccdbe$ ↓ &
$\edgeprc$ ↑ &
& 
$\delta$1 ↑ &
$\epscompdbe$ ↓ &
$\epsaccdbe$ ↓ &
$\edgeprc$ ↑ %
\\

\midrule

\multirow{3}{*}{Marigold-Depth (base)} & 
768 &
97.5 & 
5.0 & 
2.4 & 
24. & 
&
\underline{99.1} & 
6.4 & 
3.1 & 
33. %
\\

& 
1536 &
95.3 & 
4.4 & 
2.3 & 
23. & 
&
92.6 & 
5.6 & 
2.5 & 
37. %
\\

& 
3072$^{\ddag}$ &
84.1 & 
6.3 & 
2.2 & 
23. & 
&
79.3 & 
8.8	& 
7.0 & 
16. %
\\

\midrule

\multirow{2}{*}{\makecell[l]{\method{}-Depth \\ + Global cond.}} &
1536 &
97.4 & 
\underline{3.7} & 
1.9 & 
\underline{30.} & 
&
98.4 & 
4.5 & 
2.1 & 
45. %
\\

& 
3072$^{\ddag}$ &
96.7 & 
4.0 & 
\textbf{1.8} & 
30. & 
&
98.1 & 
\underline{4.4} & 
1.9 & 
\textbf{49.} %
\\

\midrule

\multirow{2}{*}{\makecell[l]{\method{}-Depth \\ + MultiDiffusion}} & 
1536 &
93.3 & 
4.0 & 
2.1 & 
25. & 
&
88.6 & 
4.5 & 
\textbf{1.6} & 
40. %
\\

& 
3072 &
82.9 & 
5.0 & 
2.0 & 
23. & 
&
82.9 & 
5.0 & 
2.0 & 
23. %
\\

\midrule

\multirow{2}{*}{Marigold-HR (best)} & 
1536 &
\textbf{98.0} & 
3.8 & 
2.0 & 
\underline{30.} & 
&
98.8 & 
4.5 & 
2.2 & 
43. %
\\

& 
3072 &
\underline{97.8} & 
\textbf{3.5} & 
\textbf{1.8} & 
\textbf{33.} & 
&
\textbf{99.2} & 
\textbf{4.1} & 
\underline{1.8} & 
\underline{48.} %
\\

\bottomrule

\end{tabular}

%% file: sec/5_conclusion.tex
\section{Conclusion}
\label{sec:conclusion}

We have presented \method{}, an affordable fine-tuning protocol for pretrained text-to-image LDMs, and a family of models for state-of-the-art image analysis tasks.
Our evaluation confirms the value of leveraging rich scene priors and diverse synthetic data across tasks such as monocular depth prediction, surface normals estimation, and intrinsic image decomposition. 
\method{} offers competitive performance across all these tasks. 
Additionally, we have presented LCM distillation and High-Resolution inference, which are adaptable to any modality. 
\method{} is trainable in under 3 GPU-days on consumer hardware, and its single-step inference has runtime comparable with other recent approaches.
List of models, web apps (spaces), code, and further reading links:\\[1em]
\begin{minipage}[t]{\linewidth}
{
\footnotesize
\begin{tabular}{@{}ll@{}}
Space Depth & \href{https://hf.co/spaces/prs-eth/marigold}{hf.co/spaces/prs-eth/marigold} \\
Space Normals & \href{https://hf.co/spaces/prs-eth/marigold-normals}{hf.co/spaces/prs-eth/marigold-normals} \\
Space Intrinsics & \href{https://hf.co/spaces/prs-eth/marigold-intrinsics}{hf.co/spaces/prs-eth/marigold-intrinsics} \\
Model Depth \texttt{v1.0} & \href{https://hf.co/prs-eth/marigold-depth-v1-0}{hf.co/prs-eth/marigold-depth-v1-0} \\
Model Depth \texttt{v1.1} & \href{https://hf.co/prs-eth/marigold-depth-v1-1}{hf.co/prs-eth/marigold-depth-v1-1} \\
Model Normals \texttt{v1.1} & \href{https://hf.co/prs-eth/marigold-normals-v1-1}{hf.co/prs-eth/marigold-normals-v1-1} \\
Model Appearance \texttt{v1.1} & \href{https://hf.co/prs-eth/marigold-iid-appearance-v1-1}{hf.co/prs-eth/marigold-iid-appearance-v1-1} \\
Model Lighting \texttt{v1.1} & \href{https://hf.co/prs-eth/marigold-iid-lighting-v1-1}{hf.co/prs-eth/marigold-iid-lighting-v1-1} \\
Model Depth-LCM \texttt{v1.0} & \href{https://hf.co/prs-eth/marigold-depth-lcm-v1-0}{hf.co/prs-eth/marigold-depth-lcm-v1-0} \\
Model Depth-HR \texttt{v1.0} & \href{https://hf.co/prs-eth/marigold-depth-hr-v1-0}{hf.co/prs-eth/marigold-depth-hr-v1-0} \\
\end{tabular}
}
\end{minipage}\\
\begin{minipage}[t]{\linewidth}
{
\footnotesize
\begin{tabular}{@{}ll@{}}
Training code & \href{https://github.com/prs-eth/Marigold}{github.com/prs-eth/Marigold} \\
Inference code & \href{https://hf.co/docs/diffusers/api/pipelines/marigold}{hf.co/docs/diffusers/api/pipelines/marigold} \\
\texttt{diffusers} tutorial & \href{https://hf.co/docs/diffusers/using-diffusers/marigold_usage}{hf.co/docs/diffusers/using-diffusers/marigold\_usage}
\end{tabular}
}
\end{minipage}

%% file: sec/ack.tex
\section*{Acknowledgements}

We thank: 
The Hugging Face team (and especially Ahsen Khaliq, Omar Sanseviero, Sayak Paul, Yiyi Xu) for 
(1) the GPU grants to host Marigold spaces and models,
(2) helping us to promote Marigold among research and content creator communities on X (Twitter), Posts, Spaces, and beyond, 
(3) guidance with integrating Marigold into the \texttt{diffusers}~\cite{von-platen-etal-2022-diffusers};
Robert Presl for creating multiple promotional 3D prints for this paper;
Alexander Becker, Dominik Narnhofer, and Xiang Zhang for discussions related to Marigold-HR;
Peter Kocsis for help with reproducing 
\cite{kocsis2024intrinsic}.

%% file: sec/bibliography.tex
\begin{IEEEbiographynophoto}{Bingxin Ke}
is a PhD student at the Photogrammetry and Remote Sensing Lab of ETH Zurich. 
His interest lies in generalizable computer vision, especially for 3D vision problems. 
He earned his Bachelor's degree in Geomatics Engineering at Wuhan University in 2020, and Master's degree in Geomatics at ETH Zurich in 2022.
\end{IEEEbiographynophoto}
\vspace{-1em}
\begin{IEEEbiographynophoto}{Kevin Qu}
is a Master student at the Photogrammetry and Remote Sensing Lab of ETH Zurich. He obtained his Bachelor's degree in Electrical and Computer Engineering at the Technical University of Munich in 2023, and is currently pursuing his Master's degree in Robotics, Systems and Control at ETH. His research interests include generative models, 3D vision and robotic perception.
\end{IEEEbiographynophoto}
\vspace{-1em}
\begin{IEEEbiographynophoto}{Tianfu Wang}
is a PhD student at the Intelligent Sensing Lab of the University of Maryland, College Park. Tianfu completed his Master's degree in Computer Science at ETH Zurich, where he worked in the Photogrammetry and Remote Sensing Lab. Prior to that, Tianfu earned his Bachelor's degree in Computer Science and Mathematics from Northwestern University. Tianfu is interested in computational imaging, generative models, and differentiable rendering.  
\end{IEEEbiographynophoto}
\vspace{-1em}
\begin{IEEEbiographynophoto}{Nando Metzger}
is a PhD student at the Photogrammetry and Remote Sensing Lab of ETH Zurich. He is interested in weakly supervised learning and super-resolution techniques and their applications to monocular depth and remote sensing. He studied Geomatics at ETH Zurich, where he received his Bachelor's degree in 2019 and his Master's degree in 2021.
\end{IEEEbiographynophoto}
\vspace{-1em}
\begin{IEEEbiographynophoto}{Shengyu Huang}
is a PhD student at the Photogrammetry and Remote Sensing Lab of ETH Zurich. He is interested in 3D vision problems and applications of neural fields beyond conventional cameras. He earned a Bachelor's degree in Surveying and Mapping Engineering at Tongji University in 2018, later he obtained his Master's degree in Geomatics from ETH Zurich in 2020.
\end{IEEEbiographynophoto}
\vspace{-1em}
\begin{IEEEbiographynophoto}{Bo Li} is a student collaborator with the Photogrammetry and Remote Sensing Lab of ETH Zurich. 
His research interests include computer graphics, generative models and efficient GPU computation for large-scale AI workloads. 
He earned dual Bachelor's degrees in Biological Science and Computer Science from Peking University and Master's degree in Computer Science at ETH Zurich.
\end{IEEEbiographynophoto}
\vspace{-1em}
\begin{IEEEbiographynophoto}{Anton Obukhov}
is an established researcher in the Photogrammetry and Remote Sensing Lab, broadly interested in computer vision and machine learning research. He joined the Computer Vision Laboratory at ETH Zurich in 2018 and received his PhD in 2022, supervised by Prof. Luc Van Gool and funded by the Toyota TRACE-Zurich project. 
He also holds a Diploma in Computational Mathematics and Cybernetics from Moscow State University, obtained in 2008.
Between these degrees, he spent a decade working in the industry: he helped NVIDIA drive the adoption of NVIDIA CUDA technology in scientific computing and later helped Ubiquiti Networks build multiple video camera products with varying degree of artificial intelligence.
\end{IEEEbiographynophoto}
\vspace{-1em}
\begin{IEEEbiographynophoto}{Konrad Schindler}
(Senior Member, IEEE) received the Diplomingenieur (MTech) degree from the Vienna University of Technology, Vienna, Austria, in 1999, and the PhD from the Graz University of Technology, Graz, Austria, in 2003. 
He was a photogrammetric engineer in the private industry and held researcher positions with the Graz University of Technology, Monash University, Melbourne, VIC, Australia, and ETH Zürich, Zürich, Switzerland.
He was an assistant professor of image understanding with TU Darmstadt, Darmstadt, Germany, in 2009. 
Since 2010, he has been a tenured professor of photogrammetry and remote sensing with ETH Zürich. 
His research interests include computer vision, photogrammetry, and remote sensing.
\end{IEEEbiographynophoto}